\documentclass[11pt]{article}

\usepackage[preprint]{acl}

\usepackage{times}
\usepackage{latexsym}
\usepackage[T1]{fontenc}
\usepackage[utf8]{inputenc}
\usepackage{microtype}
\usepackage{inconsolata}

\usepackage{amsmath}
\usepackage{amssymb}
\usepackage{booktabs}
\usepackage{multirow}
\usepackage{array}
\usepackage{graphicx}
\usepackage{subcaption}
\usepackage{enumitem}
\usepackage{xspace}

\newcommand{\atlas}{\textsc{ATLAS}\xspace}

\newcommand{\auc}{\ensuremath{\mathrm{AUC}}}
\newcommand{\atlaslite}{\textsc{ATLAS}-Lite\xspace}

\newcommand{\meituanmark}{\ensuremath{\diamondsuit}}
\newcommand{\fudanmark}{\ensuremath{\clubsuit}}
\newcommand{\atlasrunningtitle}{ATLAS: All-round Testing of Long-context Abilities across Scales}
\newcommand{\AtlasBeforeBibliography}{%
  \section*{Acknowledgments}
  We thank Zhejun Liu, Jiaqi Fan, Hang Zhou, Guangmeng Wang, Xiao Liu, Ruiqi Xu, Xing Hu, Rumei Li, Yunke Zhao, Xu Han, Shuai Liang, Huantian Lv, Dengchang Zhao, Guanyu Wu, Xurui Yang, Guoyuan Lin, and Fengcun Li for their helpful discussions and support.
}
\newcommand{\atlaslogobar}{%
  {\normalfont\normalsize
  \parbox{\textwidth}{%
    \raggedright
    \includegraphics[width=0.18\textwidth]{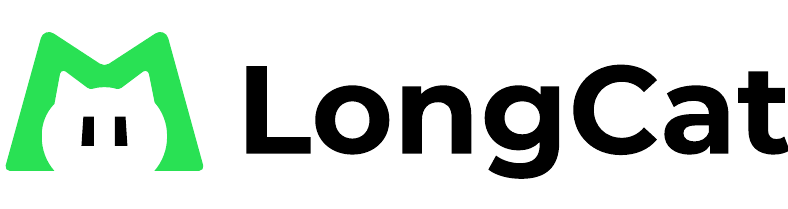}\\[-0.95em]
    \rule{\textwidth}{1.0pt}%
  }}%
}

\title{\atlaslogobar\\[2.0em]
\atlasrunningtitle}

\newcommand{\atlaspagehead}{%
  \vbox{%
    \hbox to \textwidth{%
      \hfil
      {\normalfont\footnotesize \atlasrunningtitle}%
      \hfil
      \llap{\raisebox{-0.35\height}{\includegraphics[width=0.09\textwidth]{figure/pdf/longcat-logo-full.pdf}}}%
    }%
    \vskip 2pt
    \hrule width \textwidth height 0.5pt
  }%
}
\makeatletter
\def\ps@atlasheaders{%
  \def\@oddhead{\atlaspagehead}%
  \def\@evenhead{\atlaspagehead}%
  \def\@oddfoot{\hfil\thepage\hfil}%
  \let\@evenfoot\@oddfoot
}
\makeatother

\author{
Deli Huang\textsuperscript{\meituanmark,\fudanmark,*}
\quad Cunguang Wang\textsuperscript{\meituanmark,\textdagger}
\quad Hongyin Tang\textsuperscript{\meituanmark}
\quad Zhe Tang\textsuperscript{\meituanmark}\\
Linsen Guo\textsuperscript{\meituanmark}
\quad Dongyu Ru\textsuperscript{\meituanmark}
\quad Ruoshi Yuan\textsuperscript{\meituanmark}
\quad Ziyue Zhu\textsuperscript{\meituanmark,\fudanmark,*}\\
Xiaoyu Li\textsuperscript{\meituanmark}
\quad Ziwen Wang\textsuperscript{\meituanmark}
\quad Chen Zhang\textsuperscript{\meituanmark}
\quad Anchun Gui\textsuperscript{\meituanmark}\\
Wen Zan\textsuperscript{\meituanmark}
\quad Jiaqi Zhang\textsuperscript{\meituanmark}
\quad Xuezhi Cao\textsuperscript{\meituanmark,\textdagger}
\quad Jingang Wang\textsuperscript{\meituanmark}\\
Xunliang Cai\textsuperscript{\meituanmark}
\quad Yixin Cao\textsuperscript{\fudanmark}\\[0.3em]
{\normalfont\meituanmark{} Meituan\quad \fudanmark{} Fudan University}
}

\begin{document}

\maketitle
\thispagestyle{plain}
\pagestyle{atlasheaders}
\begingroup
\renewcommand{\thefootnote}{\fnsymbol{footnote}}
\footnotetext[1]{Work done during the internship at Meituan.}
\footnotetext[2]{Correspondence: \href{mailto:wangcunguang@meituan.com}{wangcunguang@meituan.com}, \href{mailto:caoxuezhi@meituan.com}{caoxuezhi@meituan.com}.}
\endgroup

\begin{abstract}
Long-context language models now advertise context windows up to millions of tokens, yet evaluations typically report a single length or a narrow task family, masking two failure modes: performance can collapse as length grows, and strong retrieval need not transfer to downstream use. We present \atlas, a benchmarking framework that redefines long-context evaluation as length-dependent capability profiling. \atlas contributes three methodological principles: (i)~a layered taxonomy separating foundational operations from application workloads so failures can be attributed, (ii)~length-aware AUC scoring that integrates score--length curves over a fixed 8K--1M grid, replacing single-point metrics with full degradation profiles, and (iii)~ATLAScore, a harmonic-mean aggregate over taxonomy categories that penalizes imbalanced profiles, with end-to-end uncertainty propagation from subset scores through the nonlinear final aggregate. We instantiate the framework across eight capability dimensions with nine auditable components and 6,438 instances, and evaluate 26 models. Gemini-3.1-Pro-Preview leads at 128K, Claude-Opus-4.6 leads at 1M. Rankings reshuffle substantially between ATLAScore@8K-128K and ATLAScore@8K-1M: 7 models move by at least two ranks, and the two taxonomy layers share only 61\% of cross-model variance, with individual rank gaps up to 12 positions. These results support reporting long-context quality by capability and length, not by a single headline score.
\end{abstract}

\section{Introduction}
\label{sec:intro}

\begin{figure*}[t]
\centering
\includegraphics[width=0.9\textwidth]{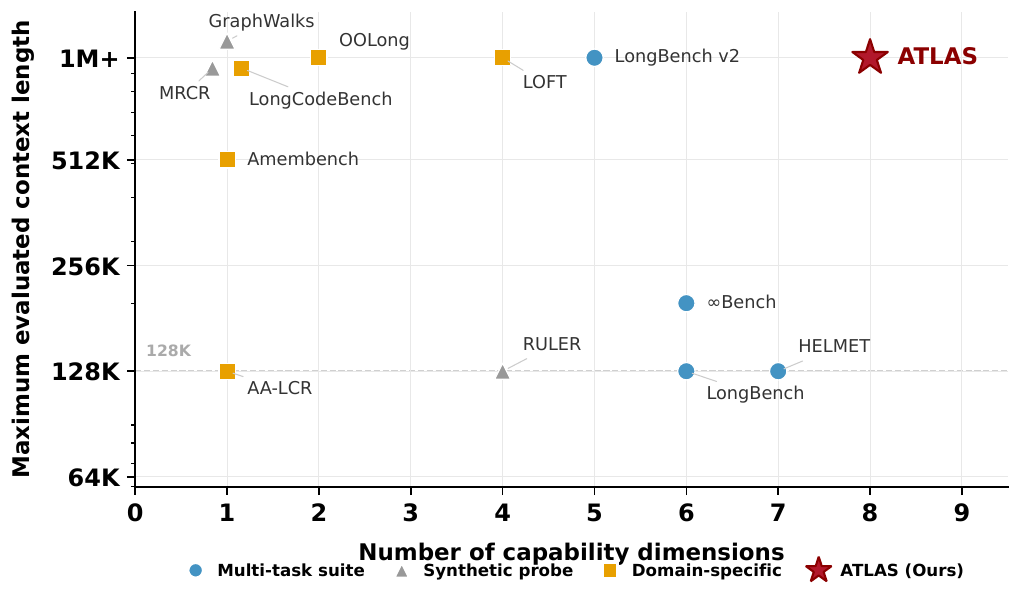}
\caption{Positioning of \atlas relative to representative long-context benchmarks by capability breadth (x-axis) and maximum evaluated context length (y-axis, log scale). Marker shapes distinguish multi-task suites, synthetic probes, and domain-specific benchmarks. Markers with identical coordinates are slightly offset for visibility. \atlas occupies the upper-right corner by combining eight capability dimensions, evaluation up to 1M tokens, and length-aware AUC scoring.}
\label{fig:overview}
\end{figure*}

As context windows expand from thousands to millions of tokens~\citep{openai2023gpt4,geminiteam2024gemini15,anthropic2024claude3}, users increasingly expect a single model call to read lengthy documents, inspect full repositories, maintain long-range conversational memory, and reason across heterogeneous evidence. The advertised maximum window, however, is not the same as effective long-context capability. Models may perform well at moderate lengths while degrading sharply near their nominal maximum, and success on retrieval-style probes may not transfer to application workloads such as in-context learning, code understanding, or cross-document analysis.

Existing evaluation does not yet capture these distinctions systematically. Synthetic probes such as needle-in-a-haystack, MRCR, BABILong, and GraphWalks isolate controlled operations but cover only a narrow slice of model behavior~\citep{kamradt2023niah,openai2025mrcr,kuratov2024babilong,openai2025graphwalks}. Long-text and multitask suites such as SCROLLS, ZeroSCROLLS, L-Eval, LongBench, and HELMET broaden task coverage, but they either operate mostly at moderate lengths or stop at 128K~\citep{shaham2022scrolls,shaham2023zeroscrolls,an2023leval,bai-etal-2024-longbench,yen2025helmet}. Ultra-long benchmarks such as $\infty$Bench, LOFT, and LongCodeBench reach larger scales, but they do not provide a unified, length-aware diagnostic profile across capability dimensions~\citep{zhang2024inftybench,Lee2024LongContext,rando2025longcodebench}. Figure~\ref{fig:overview} summarizes this gap: current benchmarks tend to be either broad but short, or long but narrow.

We introduce \atlas, a benchmarking framework that evaluates long-context language models as length-dependent capability profiles rather than single-point scores. \atlas is built around a simple premise: a reliable long-context benchmark should answer both \emph{what} a model can do and \emph{how long} it can continue doing it. To this end, it decomposes long-context ability into a layered capability taxonomy, evaluates each dimension across a geometric length grid, and summarizes performance with curve-integrated scores that separate medium-length quality from ultra-long robustness.

The empirical message is direct: single-window evaluation can change the apparent model conclusion. Seven of 26 models shift by two or more ranks when the reporting scope extends from 128K to 1M---including one frontier model that drops four positions---demonstrating that medium-length quality and ultra-long robustness are distinct signals that a framework must report separately. The taxonomy also matters: foundational and application aggregates share only 61\% of cross-model variance at 128K, with individual models shifting by up to 12 positions between the two layer rankings, confirming that no single task dimension suffices.

We make three contributions:
\begin{enumerate}[leftmargin=1.2em]
 \item We formalize a layered evaluation framework for long-context models, defining \emph{what} must be measured (a 3+5 capability taxonomy with empirically validated non-redundancy) and \emph{how} it should be scored. Instead of selecting a single operating length, \atlas integrates each dimension's score--length curve over a fixed 8K--1M grid via length-aware AUC, and aggregates across categories with a harmonic mean (ATLAScore) that penalizes imbalanced profiles. The framework is component-agnostic: individual benchmarks serve as replaceable instantiations of each dimension, selected by explicit criteria (length extensibility, deterministic scoring, cross-model discrimination) and validated through leave-one-out stability analysis.
 \item We provide, to our knowledge, the first end-to-end confidence-interval propagation pipeline for long-context evaluation, propagating uncertainty through AUC aggregation and nonlinear harmonic ATLAScore computation, with Monte Carlo validation in Appendix~\ref{app:ci}.
 \item We instantiate the framework's eight dimensions with nine components spanning 6,438 instances, evaluate 26 models, and show that single-window rankings obscure length-dependent degradation and mask non-redundant capability differences between the two taxonomy layers.
\end{enumerate}

\section{What Long-Context Evaluation Must Measure}
\label{sec:background}

Long-context evaluation must separate three quantities that are often conflated: how much context a model accepts, which long-context operations it can perform, and how those operations degrade as input length grows. A single score at 128K cannot distinguish a model that remains stable through 1M from one that reaches a cliff immediately after 128K. Prior work further shows that effective context can be shorter than nominal context and that input length, evidence position, evidence dispersion, and task scope affect success~\citep{liu2023lostmiddlelanguagemodels,levy2024sametask,hsieh2024ruler,goldman2024isitreally}. A benchmark therefore needs controlled length slices and degradation curves, not only endpoint scores or advertised window sizes.

The remaining gap is not a shortage of long-context tasks---recent surveys already document a large benchmark and modeling landscape~\citep{liu2025comprehensivesurveylongcontext}---but the absence of a principled benchmarking framework that turns task-level observations into comparable, uncertainty-quantified capability profiles. Retrieval, aggregation, multi-step reasoning, in-context learning, code understanding, memory, and cross-document analysis can diverge sharply, as prior work shows for synthetic recall versus downstream performance~\citep{yen2025helmet} and for aggregation versus retrieval~\citep{bertsch2025oolongevaluatinglongcontext}. Breadth without length-aware scoring produces task-level snapshots rather than degradation profiles, and length without breadth reveals only one bottleneck. Filling this gap requires three properties: a capability decomposition so that failures can be attributed rather than averaged away, a fixed length grid so that degradation is measured rather than assumed, and a compact instance budget so that the full protocol remains repeatable for model iteration. \atlas is designed around these three properties; its specific component choices are validated in Section~\ref{sec:construction}.

\section{The \atlas Benchmark}
\label{sec:method}

\subsection{Capability Taxonomy}
\label{sec:taxonomy}

\atlas specifies \emph{dimensions} and \emph{scoring semantics} independently of the benchmarks that fill each slot; individual components are replaceable instantiations selected and validated in Section~\ref{sec:construction}. The taxonomy organizes evaluation into a layered 3+5 structure. The foundational layer comprises three competence probes (retrieval, aggregation, and multi-step reasoning) whose answers are fully determined by the input, making failures easy to attribute. The application layer covers five realistic workloads (question answering, in-context learning, code understanding, long-range memory, and holistic assessment) that additionally require domain knowledge, instruction following, or conversational grounding. The two layers can disagree: a model may retrieve evidence reliably yet fail on downstream use, or vice versa. Reporting them separately therefore yields a diagnostic profile rather than a single scalar; Section~\ref{sec:layer_validation} tests whether this separation produces non-redundant empirical signals.

Within the application layer, the holistic assessment dimension is scored separately because its tasks cannot be meaningfully length-sliced---they do not admit controlled length extension without distorting the task itself. Rather than excluding these signals, \atlas evaluates them at their original instance lengths. The scoring pipeline therefore operates over three categories---foundational, length-sliced application, and holistic assessment---combined via harmonic aggregation (Section~\ref{sec:scoring}). This keeps the framework extensible: future tasks that resist length slicing can enter the holistic category without disrupting the AUC-based scoring of the other two.

\begin{table*}[t]
\caption{The eight \atlas capability dimensions and their current component instantiations. The first seven dimensions are evaluated over length slices and scored via AUC; the holistic assessment dimension is evaluated at original instance lengths.}
\label{tab:components}
\centering
\footnotesize
\renewcommand{\arraystretch}{1.18}
\resizebox{\textwidth}{!}{
\begin{tabular}{llllll}
\toprule
Layer & Dimension & Component & Range & Metric & Instances \\
\midrule
\multirow{3}{*}{Foundational}
& Retrieval & MRCR-8 needle & 8K--1M & EM & 792 \\
& Aggregation & OOLong-Synth & 8K--1M & Answer-level & 800 \\
& Reasoning & GraphWalks Extend & 8K--1M & F1 & 800 \\
\midrule
\multirow{5}{*}{Application}
& QA & LOFT-Text Retrieval Extend & 8K--1M & MRecall@K & 800 \\
& ICL & Helmet-ICL Extend & 8K--1M & Acc & 800 \\
& Code & LongCodeBench & 32K--1M & Acc / Pass@1 & 1,043 \\
& Memory & AMemBench-ACU & 8K--1M & QPEM & 800 \\
& Holistic assessment & LongBench-v2 / AA-LCR & varied & Acc & 603 \\
\midrule
\multicolumn{5}{l}{\textbf{Total}} & \textbf{6,438} \\
\bottomrule
\end{tabular}
}
\end{table*}

\subsection{Length-Aware Scoring}
\label{sec:scoring}

Length-aware scoring is a methodological component of \atlas, not only a leaderboard convention: it defines how discrete length slices become a comparable estimate of operating robustness. We evaluate each dimension over a fixed set of representative context lengths
\[
\mathcal{L} = \{8\mathrm{K}, 16\mathrm{K}, 32\mathrm{K}, 64\mathrm{K}, 128\mathrm{K}, \ldots, 1\mathrm{M}\},
\]
spanning common deployment lengths and frontier ultra-long regimes. The slices are geometrically spaced, so that the benchmark observes both gradual degradation and abrupt cliffs without requiring dense measurement at every possible context length.

For a reporting scope $L^\star \in \{128\mathrm{K},1\mathrm{M}\}$, let $\mathcal{L}_{\leq L^\star} = \{\ell_0, \ell_1, \ldots, \ell_n\}$ denote the ordered slices up to that scope. For dimension $d$, let $s_d(\ell) \in [0,100]$ be the normalized score at length $\ell$. Following the principle that a single-point measurement cannot capture how performance degrades across a continuous axis~\citep{zhou2025gsminfinite}, we adopt a normalized trapezoidal AUC score. With $\Delta_i=\ell_{i+1}-\ell_i$,
\[
\auc_d(L^\star) =
\frac{\sum_{i=0}^{n-1}\Delta_i\,[s_d(\ell_i) + s_d(\ell_{i+1})]/2}{\ell_n - \ell_0}.
\]
Because the slices grow geometrically (each doubling), the trapezoidal rule naturally gives more weight to longer-range intervals where degradation is most consequential. ATLAScore@8K-128K and ATLAScore@8K-1M are cumulative summaries over all slices up to 128K and 1M, not scores measured only at exactly those lengths. Appendix~\ref{app:weights} shows that rankings are stable under alternative length weights.

Let $\mathcal{D}_{\mathrm{base}}$ denote the three foundational dimensions, $\mathcal{D}_{\mathrm{app}}$ the four length-sliced application dimensions, and $\mathcal{D}_{\mathrm{hol}}$ the holistic assessment dimensions. We compute three category aggregates:
\begin{align*}
B(L^\star) &=
\frac{1}{|\mathcal{D}_{\mathrm{base}}|}
\sum_{d \in \mathcal{D}_{\mathrm{base}}} \auc_d(L^\star), \\
C(L^\star) &=
\frac{1}{|\mathcal{D}_{\mathrm{app}}|}
\sum_{d \in \mathcal{D}_{\mathrm{app}}} \auc_d(L^\star), \\
S(L^\star) &=
\frac{1}{|\mathcal{D}_{\mathrm{hol}}|}
\sum_{d \in \mathcal{D}_{\mathrm{hol}}} s_d,
\end{align*}
where $S$ is fixed across reporting scopes because the holistic assessment dimensions are evaluated at their original instance lengths rather than over length slices. The overall score is the harmonic mean of the three category aggregates:
\[
\begin{aligned}
\mathrm{ATLAScore}(L^\star)
&= 3\Bigl(
B(L^\star)^{-1} + C(L^\star)^{-1} \\
&\quad + S(L^\star)^{-1}
\Bigr)^{-1}.
\end{aligned}
\]
The harmonic mean prevents a model from receiving a high overall score by excelling in only one category; Appendix~\ref{app:aggregation} shows that relative rankings are nevertheless robust to alternative aggregation rules. Appendix~\ref{app:holistic_sensitivity} further tests whether the holistic category's composition or weight materially changes the ranking.

\begin{figure*}[t]
\centering
\begin{subfigure}[b]{0.52\textwidth}
    \includegraphics[width=\linewidth]{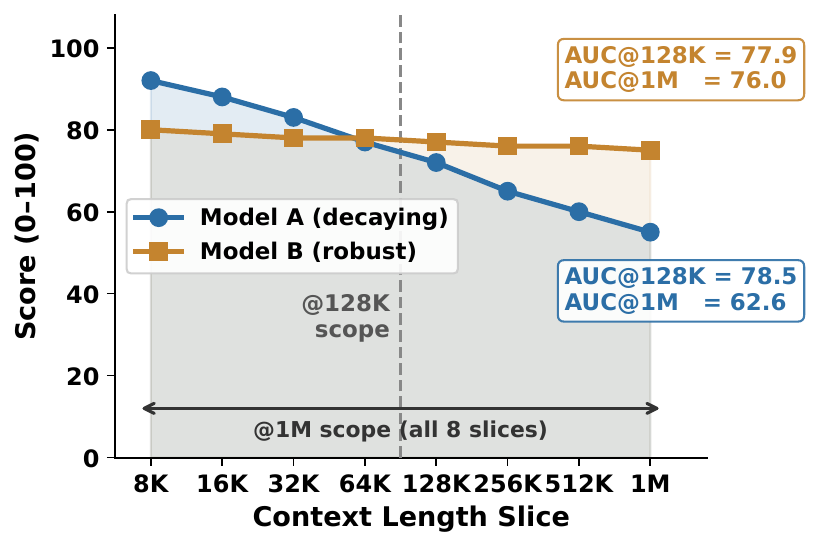}
    \caption{Length-aware \auc}
    \label{fig:schematic_auc}
\end{subfigure}
\hfill
\begin{subfigure}[b]{0.46\textwidth}
    \includegraphics[width=\linewidth]{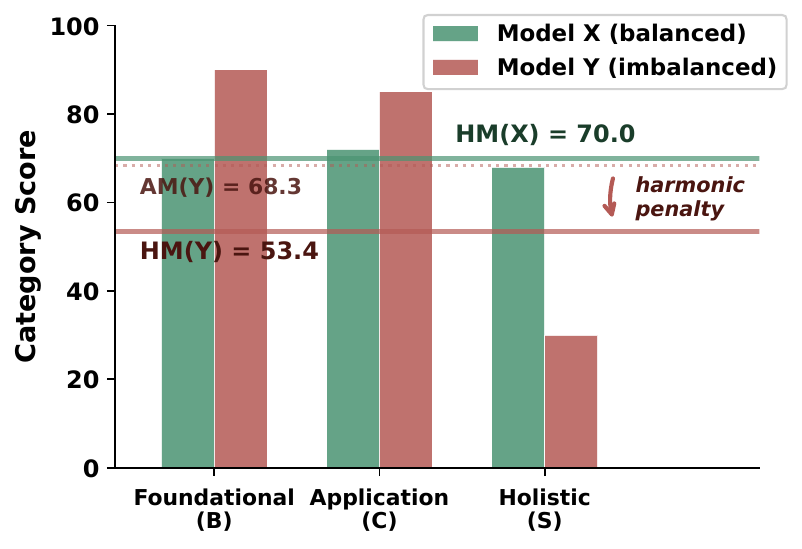}
    \caption{Harmonic ATLAScore}
    \label{fig:schematic_hm}
\end{subfigure}
\caption{Geometric intuition for the \atlas scoring pipeline. (a)~Length-aware \auc{} aggregates scores across length slices up to the reporting scope. (b)~Harmonic mean (HM) penalizes category imbalance relative to the arithmetic mean (AM).}
\label{fig:scoring_schematic}
\end{figure*}

Figure~\ref{fig:scoring_schematic} illustrates this design. In panel~(a), the cumulative \auc{} captures a model's score profile across length slices rather than a single-point measurement, so a model that starts strong but degrades yields a lower \auc{} than one with modest but stable performance. In panel~(b), the harmonic mean is pulled toward the weakest category, so a model with one deficient dimension scores substantially lower than a balanced competitor even when their arithmetic means coincide.

The two reporting scopes serve different interpretation needs. ATLAScore@8K-128K summarizes the regime most current production systems can plausibly support, while ATLAScore@8K-1M stresses frontier ultra-long use. Comparing the two is itself informative: a model that ranks highly at 128K but drops at 1M has a different deployment profile from a model with slightly lower short-scope quality but stronger length robustness.

We also propagate statistical uncertainty end to end through the scoring pipeline. Subset-level confidence intervals are computed with task-appropriate estimators, including cluster correction when multiple instances share a base context and weighted combination for composite components. Since AUC is a weighted average over independent length slices, its variance propagates linearly. The final harmonic ATLAScore is nonlinear, so we use the delta method to propagate category-level variances into the reported confidence intervals. Appendix~\ref{app:ci} provides the derivation and a Monte Carlo validation.

\subsection{Component Selection and Validation}
\label{sec:construction}

Table~\ref{tab:components} lists the nine components that currently instantiate the eight taxonomy dimensions. Each framework dimension is filled by a component satisfying three necessary criteria: length extensibility, deterministic scoring, and cross-model discrimination. Components are replaceable: leave-one-out ablation preserves the full ranking, and future components meeting the same criteria can be substituted without changing the scoring pipeline. Open-ended summarization is excluded from the current release because evaluator-independent scoring for long summaries remains unreliable: n-gram overlap metrics are weak proxies for summary quality, while model-based judging introduces judge-version dependence~\citep{lin2004rouge,fabbri2021summeval,zheng2024judging}.

Several components required controlled length extension. GraphWalks officially provides 128K and 1M slices; we synthesize intermediate slices using the official generation procedure. HELMET-ICL supports configurable generation through public scripts, and LOFT-Text Retrieval adapts LOFT's text-retrieval instances into a HELMET-style long-context retrieval format, with context length controlled by padding to fill all eight slices. The assistant conversation understanding version of AMemBench (AMemBench-ACU) converts generated long-horizon interaction transcripts into static atomic-content-unit extraction instances. LongCodeBench starts at 32K because repository-scale inputs cannot be meaningfully compressed to 8K or 16K; its unsupported 8K and 16K slices are omitted from per-slice application aggregation. The holistic assessment category is currently instantiated by LongBench-v2, which evaluates realistic long-context multitask reasoning, and AA-LCR, which evaluates cross-document analytical reasoning, both at their original instance lengths. Appendix~\ref{app:component_details} describes each component and extension procedure.

The resulting set of dimensions is empirically broad rather than redundant. At 128K, the seven length-sliced dimensions have mean pairwise Spearman correlation $\bar{\rho}=0.64$ across 26 models, with 6 of 21 unrounded pairs below $\rho=0.50$. Code is the most independent dimension: its correlations with Retrieval ($\rho=0.07$) and Reasoning ($\rho=0.17$) are not statistically significant. Each dimension also separates models: cross-model standard deviation is at least 8.9 at 128K, with Retrieval and ICL among the most discriminative dimensions. A leave-one-dimension-out analysis shows that removing any single length-sliced dimension preserves the full ranking with Spearman $\rho \geq 0.97$ at 128K and $\rho \geq 0.96$ at 1M, indicating that no single dimension dominates the score. Appendix~\ref{app:component_validation} provides the full correlation, discriminability, and leave-one-dimension-out analyses.

\atlas is compact by design: 6,438 total instances cover all dimensions and length slices, which keeps the full protocol repeatable for model iteration. For faster development loops, Appendix~\ref{app:atlas_lite} evaluates a reduced \atlaslite screening protocol and clarifies why it should not replace full ATLAScore reporting.

\section{Empirical Study}
\label{sec:experiments}

\subsection{Evaluation Setup}

We evaluate 26 proprietary and open-weight long-context language models over the eight \atlas length slices. We use each provider's recommended inference settings to reflect realistic deployment conditions and keep task instructions and scoring semantics fixed across models. If a model's advertised context length falls below a target slice, the evaluation harness applies middle truncation and the score reflects the model's actual capability under its context-length constraint, following prior long-context evaluation practice~\citep{bai2024longbenchv2}.

Full category leaderboards and confidence intervals are in Appendix~\ref{app:leaderboard}; the main text focuses on the empirical claims needed to validate the benchmark design. Specifically, the experiments address three questions:
(1)~comparing the 128K and 1M reporting scopes tests whether length-aware scoring changes model conclusions;
(2)~analyzing dimension-level decay patterns tests whether long-context degradation is capability-specific;
(3)~contrasting foundational and application scores tests whether the layered taxonomy captures non-redundant signals.
For cost-constrained screening, Appendix~\ref{app:atlas_lite} analyzes \atlaslite as a 128K-only development protocol rather than a substitute for full reporting.

Some main-text figures use representative model subsets for readability, and each caption states the displayed scope.

\subsection{Length-Aware Scoring Changes Model Rankings}
\label{sec:ranking}

\begin{figure*}[t]
\centering
\includegraphics[width=0.60\textwidth]{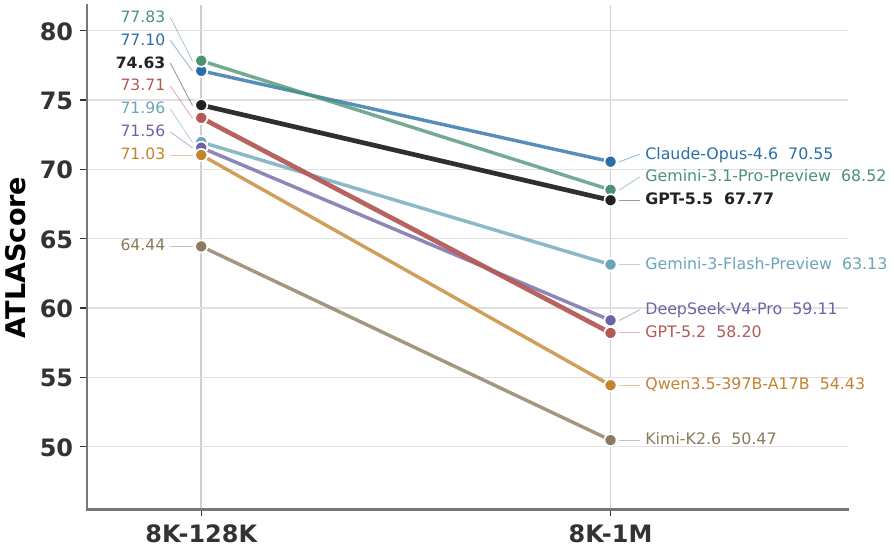}
\caption{ATLAScore comparison between the 128K and 1M reporting scopes for eight representative models (frontier leaders, a notable-drop case, and open-weight representatives). Full 26-model results in Appendix~\ref{app:leaderboard}.}
\label{fig:selected_models_scope}
\end{figure*}

Figure~\ref{fig:selected_models_scope} compares ATLAScore@8K-128K with ATLAScore@8K-1M for eight representative models. Across the full set, rankings change substantially when the reporting scope extends to the ultra-long regime: 20 of 26 models shift position, with 7 moving by two or more ranks. The upper tier changes order: Gemini-3.1-Pro-Preview leads at 128K, Claude-Opus-4.6 leads at 1M, and GPT-5.5 remains third in both scopes. GPT-5.2 drops from 4th to 8th---losing over 15 points---because its degradation is far steeper than that of the 1M leaders. Conversely, Gemini-3-Flash-Preview rises from 7th to 5th because its longer-context degradation is milder than several models ranked above it at 128K. Among open-weight models, DeepSeek-V4-Pro remains closer to the frontier group than Qwen3.5-397B-A17B and Kimi-K2.6 at 1M.

To quantify this variation, we define relative decay as $(\text{ATLAScore@8K-128K} - \text{ATLAScore@8K-1M})\,/\,\text{ATLAScore@8K-128K}$. The mean relative decay across all 26 models is 24.3\%, but the spread is large. Claude-Opus-4.6 has the lowest decay at 8.5\%, while GLM-4.7 (Non-reasoning) loses 60.5\%. This variation is precisely what single-scope reporting hides: a model's rank at 128K does not reliably predict its rank once the operating regime approaches 1M. These rank changes are not artifacts of noise: 95\% confidence intervals on ATLAScore (Tables~\ref{tab:leaderboard_128k}--\ref{tab:leaderboard_1m}) are typically 1--2 points, well below the 5--16 point gaps driving the reshuffling. Appendix~\ref{app:rank_migration} visualizes the full rank migration pattern.

\subsection{Long-Context Degradation Is Capability-Specific}
\label{sec:decay}

\begin{figure*}[t]
\centering
\includegraphics[width=\textwidth]{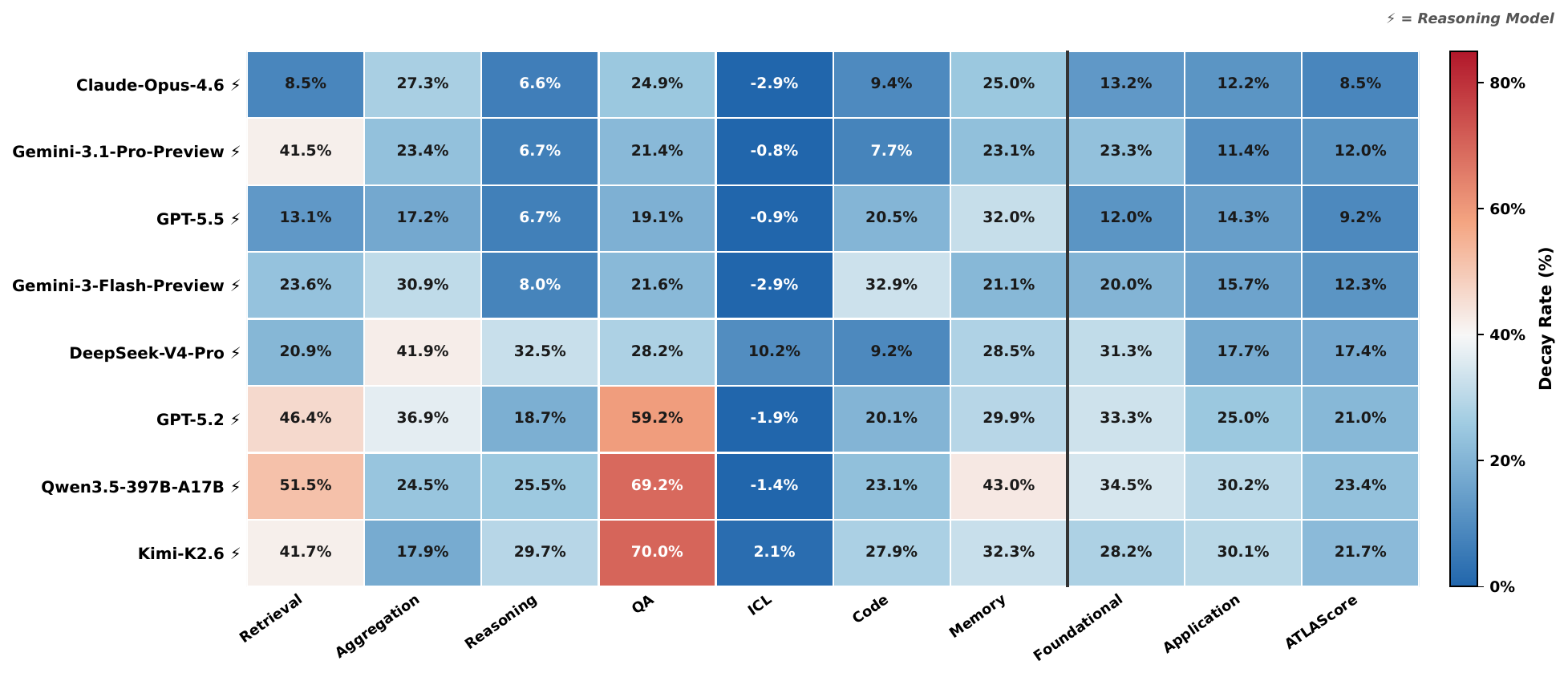}
\caption{Capability decay from the 128K to 1M reporting scope for eight
representative models, decomposed by seven length-sliced dimensions and three
aggregate scores. Models are ordered by ATLAScore@8K-1M (highest at top).
Blue = low decay; red = severe degradation. The vertical divider separates
dimension-level scores from aggregate scores. The full 26-model heatmap is reported in Appendix~\ref{app:decay}.}
\label{fig:decay_heatmap}
\end{figure*}

The rank reshuffling is caused by heterogeneous dimension-level decay, not uniform score shrinkage. Figure~\ref{fig:decay_heatmap} decomposes the relative drop from the 128K scope to the 1M scope for the representative models across capability dimensions and aggregate scores. For dimension columns, decay is measured as $(\text{AUC@8K-128K} - \text{AUC@8K-1M})\,/\,\text{AUC@8K-128K}$; for aggregate columns, the same ratio is applied to the corresponding aggregate score. This isolates each capability's degradation independently of the aggregate ATLAScore reported in Section~\ref{sec:ranking}. Retrieval and QA are the most decay-prone dimensions: multiple models lose more than 40\% of their 128K score when the evaluation extends to 1M, with extreme cases exceeding 70\%. By contrast, Code understanding shows lower and more uniform decay for many models, suggesting that repository-scale code tasks stress a different bottleneck than evidence retrieval over unstructured text.

Within-model variation is often as important as cross-model variation. GPT-5.2 remains extremely strong on ICL at 1M (97.96) but drops sharply on Retrieval (decay 46.4\% from 88.21 to 47.31) and QA (decay 59.2\% from 74.87 to 30.51), explaining its fall from 4th to 8th overall. The full decay analysis shows similar asymmetries elsewhere: Gemini-3.1-Pro-Preview retains high ICL performance (98.32), but its Retrieval score falls 41.5\% from 84.96 to 49.67. Appendix~\ref{app:length_curves} complements this ratio-based view with per-slice ATLAScore trajectories. These profiles indicate that "long-context degradation" is not a single failure mode. A model may preserve in-context learning while losing evidence retrieval, or retain structured code behavior while failing open-document QA.

A descriptive pattern emerges when grouping by inference configuration. We classify a model as a \emph{reasoning model} if its officially recommended setting produces explicit chain-of-thought (CoT) traces~\citep{wei2022chainofthought} during inference---i.e., models equipped with an extended ``thinking'' mode. Under this criterion, reasoning models exhibit systematically lower overall decay than non-reasoning models. Eleven of fifteen reasoning models have ATLAScore decay below 25\%, whereas non-reasoning models span a much wider range (19.8\% to 60.5\%). This is a correlation under deployed settings rather than a causal attribution: disentangling the roles of extended generation budgets, reinforcement-learning-based training, and data composition requires controlled ablations beyond the scope of this benchmark study.

\subsection{The Layered Taxonomy Captures Non-Redundant Signals}
\label{sec:layer_validation}

\begin{figure*}[t]
\centering
\begin{subfigure}[b]{0.48\textwidth}
    \centering
    \includegraphics[width=\linewidth]{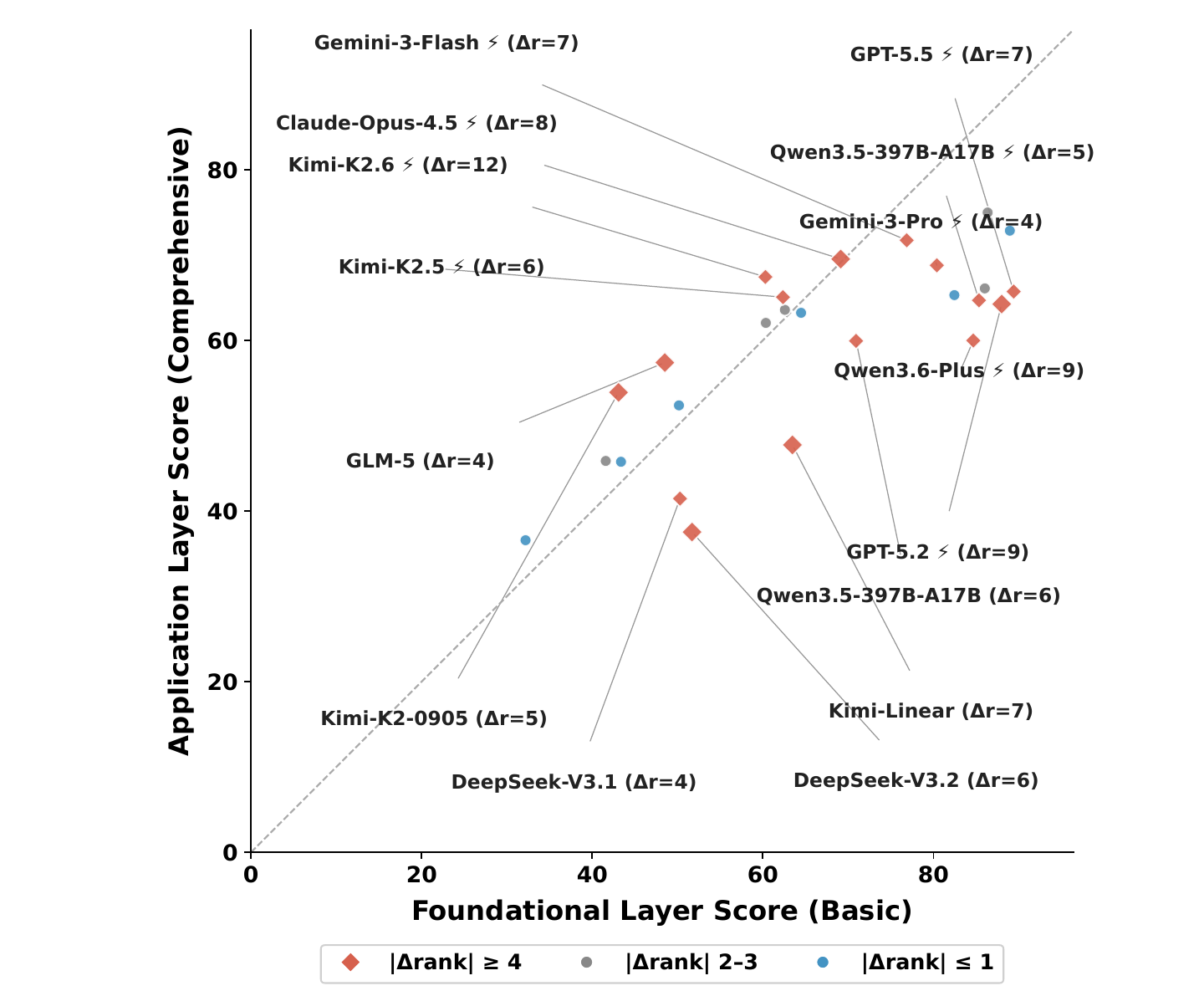}
    \caption{128K Context}
\end{subfigure}
\hfill
\begin{subfigure}[b]{0.48\textwidth}
    \centering
    \includegraphics[width=\linewidth]{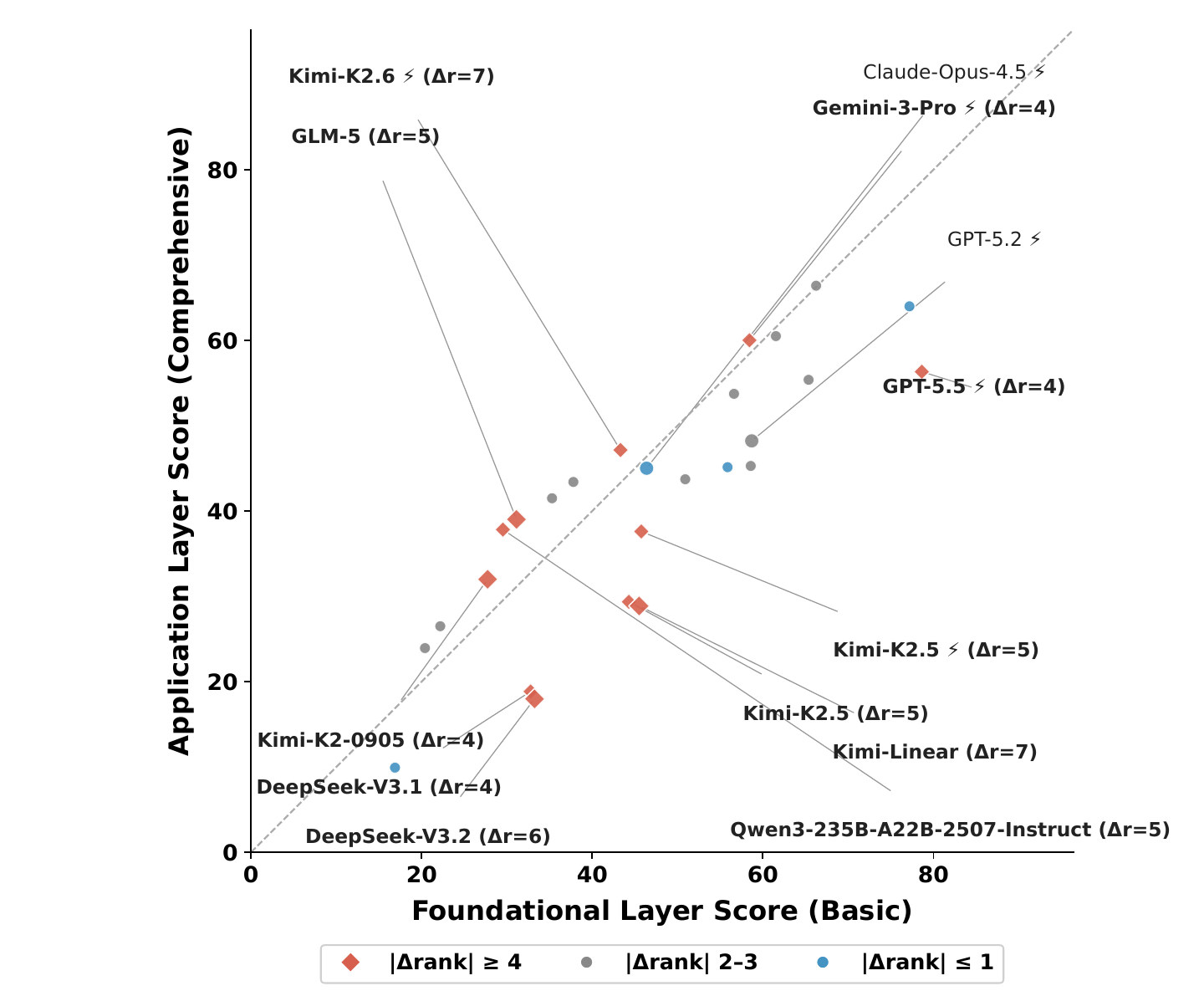}
    \caption{1M Context}
\end{subfigure}
\caption{Foundational--application layer scores for all 26 models at (a)~128K and (b)~1M. The dashed line marks score parity. Diamond markers highlight models with $|\Delta\text{rank}| = |\mathrm{rank}_B-\mathrm{rank}_C| \geq 4$, the absolute gap between foundational-layer and application-layer ranks within the same reporting scope; selected point labels abbreviate this gap as $\Delta r$.}
\label{fig:rank_discrepancy}
\end{figure*}

The foundational and application layers are designed to capture non-redundant signals. Prior work shows that synthetic retrieval performance can correlate poorly with downstream long-context task performance~\citep{yen2025helmet,bertsch2025oolongevaluatinglongcontext}; we test whether this holds in \atlas by comparing the foundational aggregate $B(L^\star)$ with the application aggregate $C(L^\star)$ (defined in Section~\ref{sec:scoring}) across all 26 models. The original-length holistic assessment category $S$ is excluded from this layer comparison because it is scored outside the \atlas length grid. At 128K, $B$ and $C$ have Pearson $R^2 = 0.61$ and Spearman $\rho = 0.74$; at 1M, they have $R^2 = 0.73$ and $\rho = 0.88$. The positive correlation is expected, but the unexplained variance is large enough to change model interpretation.

Figure~\ref{fig:rank_discrepancy} shows these discrepancies: points far from the parity line have mismatched foundational and application scores, while diamond markers identify models whose layer-rank gap is large enough to change qualitative interpretation. At 128K, 15 models shift by at least four positions between the foundational and application layer rankings, and the maximum shift is 12 positions. Kimi-K2.6 ranks 18th on foundational tasks but 6th on application tasks, while GPT-5.2 ranks 3rd on foundational tasks but 12th on length-sliced application tasks. Qwen3.6-Plus shows a similar foundational-heavy imbalance, ranking 7th on the foundational layer but 16th on applications. The discrepancy persists at 1M: although the two aggregates are more aligned ($R^2 = 0.73$, $\rho = 0.88$), 11 of 26 models still have $|\Delta\text{rank}| \geq 4$, with a maximum shift of 7 positions. These are not small presentation differences; they identify distinct failure modes that a single composite score would obscure.

Dimension-level profiles make these discrepancies actionable. A deployment that depends on multi-reference retrieval-augmented generation (RAG)~\citep{lewis2020rag} should not treat high ICL or holistic assessment scores as a substitute for retrieval robustness. A repository assistant should not over-weight general QA strength if the Code dimension is weak. Conversely, a demonstration-heavy classification workflow may prefer a model with high ICL even if it is not the strongest synthetic retriever. Appendix~\ref{app:case_studies} decomposes representative layer gaps at 128K, and Appendix~\ref{app:radar} extends the profile view with radar charts for all 26 models at both reporting scopes.

\section{Conclusion}
\label{sec:conclusion}

We presented \atlas, a length-aware and multi-dimensional benchmark framework for long-context language models. The key design choice is to treat long-context quality as a profile over both capability dimensions and context lengths, rather than as a single score at an advertised window size. Across 26 models, this design changes model interpretation: rankings reshuffle between 128K and 1M, degradation is highly capability-specific, and foundational strength explains only part of length-sliced application performance. These results support a stricter reporting norm for long-context models: public comparisons should state both the evaluated length scope and the capability profile behind the aggregate score.

\section*{Limitations}

The current \atlas release is bounded in scope: it is English-only; holistic assessment uses original benchmark lengths; AMemBench-ACU relies on model-generated transcripts; LongCodeBench begins at 32K; and open-ended summarization is omitted because evaluator-independent scoring remains unreliable. Public benchmark components also require periodic refresh to reduce contamination risk~\citep{jacovi2023contamination}. Future work will add multilingual and open-ended generation coverage, stronger refresh mechanisms, and analyses of training choices for ultra-long robustness. We will release the data, scoring scripts, and construction code upon publication.

\section*{Ethical considerations}

\atlas is intended to improve transparency in long-context model evaluation rather than to promote a single universal ranking. Because benchmark scores can influence model selection and deployment decisions, we report both aggregate scores and capability-level profiles to reduce the risk of overclaiming from a single number. The benchmark uses public or constructed evaluation components and avoids collecting private user data. For model-generated AMemBench-ACU transcripts, we treat generation bias as both a methodological and ethical concern: such data may favor models similar to the generators, so we document the construction procedure and identify this limitation explicitly. Public releases should preserve dataset licenses and avoid exposing proprietary model outputs beyond what is permitted by the corresponding providers. As with other benchmarks, \atlas may be misused for leaderboard optimization or selective reporting; we therefore recommend reporting the evaluated length scope, component scores, and known limitations alongside any headline ATLAScore.

\ifdefined\AtlasBeforeBibliography
\AtlasBeforeBibliography
\fi

\bibliography{custom}

\begin{thebibliography}{32}
\providecommand{\natexlab}[1]{#1}

\bibitem[{An et~al.(2024)An, Gong, Zhong, Zhao, Li, Zhang, Kong, and
  Qiu}]{an2023leval}
Chenxin An, Shansan Gong, Ming Zhong, Xingjian Zhao, Mukai Li, Jun Zhang,
  Lingpeng Kong, and Xipeng Qiu. 2024.
\newblock \href {https://doi.org/10.18653/v1/2024.acl-long.776} {L-eval:
  Instituting standardized evaluation for long context language models}.
\newblock In \emph{Proceedings of the 62nd Annual Meeting of the Association
  for Computational Linguistics (Volume 1: Long Papers)}, pages 14388--14411.

\bibitem[{{Anthropic}(2024)}]{anthropic2024claude3}
{Anthropic}. 2024.
\newblock \href
  {https://www-cdn.anthropic.com/de8ba9b01c9ab7cbabf5c33b80b7bbc618857627/Model_Card_Claude_3.pdf}
  {The claude 3 model family: Opus, sonnet, haiku}.
\newblock Model card.

\bibitem[{{Artificial Analysis}(2025)}]{artificialanalysis2025lcr}
{Artificial Analysis}. 2025.
\newblock \href {https://artificialanalysis.ai/articles/announcing-aa-lcr}
  {Announcing artificial analysis long context reasoning ({AA-LCR})}.
\newblock Artificial Analysis article.

\bibitem[{Bai et~al.(2024)Bai, Lv, Zhang, Lyu, Tang, Huang, Du, Liu, Zeng, Hou,
  Dong, Tang, and Li}]{bai-etal-2024-longbench}
Yushi Bai, Xin Lv, Jiajie Zhang, Hongchang Lyu, Jiankai Tang, Zhidian Huang,
  Zhengxiao Du, Xiao Liu, Aohan Zeng, Lei Hou, Yuxiao Dong, Jie Tang, and
  Juanzi Li. 2024.
\newblock \href {https://doi.org/10.18653/v1/2024.acl-long.172} {{L}ong{B}ench:
  A bilingual, multitask benchmark for long context understanding}.
\newblock In \emph{Proceedings of the 62nd Annual Meeting of the Association
  for Computational Linguistics (Volume 1: Long Papers)}, pages 3119--3137,
  Bangkok, Thailand. Association for Computational Linguistics.

\bibitem[{Bai et~al.(2025)Bai, Tu, Zhang, Peng, Wang, Lv, Cao, Xu, Hou, Dong,
  Tang, and Li}]{bai2024longbenchv2}
Yushi Bai, Shangqing Tu, Jiajie Zhang, Hao Peng, Xiaozhi Wang, Xin Lv, Shulin
  Cao, Jiazheng Xu, Lei Hou, Yuxiao Dong, Jie Tang, and Juanzi Li. 2025.
\newblock \href {https://doi.org/10.18653/v1/2025.acl-long.183} {{L}ong{B}ench
  v2: Towards deeper understanding and reasoning on realistic long-context
  multitasks}.
\newblock In \emph{Proceedings of the 63rd Annual Meeting of the Association
  for Computational Linguistics (Volume 1: Long Papers)}, pages 3639--3664,
  Vienna, Austria. Association for Computational Linguistics.

\bibitem[{Bertsch et~al.(2025)Bertsch, Pratapa, Mitamura, Neubig, and
  Gormley}]{bertsch2025oolongevaluatinglongcontext}
Amanda Bertsch, Adithya Pratapa, Teruko Mitamura, Graham Neubig, and Matthew~R.
  Gormley. 2025.
\newblock \href {https://arxiv.org/abs/2511.02817} {Oolong: Evaluating long
  context reasoning and aggregation capabilities}.
\newblock \emph{Preprint}, arXiv:2511.02817.

\bibitem[{Fabbri et~al.(2021)Fabbri, Kryściński, McCann, Xiong, Socher, and
  Radev}]{fabbri2021summeval}
Alexander~R. Fabbri, Wojciech Kryściński, Bryan McCann, Caiming Xiong,
  Richard Socher, and Dragomir Radev. 2021.
\newblock \href {https://doi.org/10.1162/tacl_a_00373} {Summeval: Re-evaluating
  summarization evaluation}.
\newblock \emph{Transactions of the Association for Computational Linguistics},
  9:391--409.

\bibitem[{{Gemini Team} et~al.(2024){Gemini Team}, Georgiev, Lei, Burnell, Bai,
  Gulati et~al.}]{geminiteam2024gemini15}
{Gemini Team}, Petko Georgiev, Ving~Ian Lei, Ryan Burnell, Libin Bai, Anmol
  Gulati, and 1 others. 2024.
\newblock \href {https://arxiv.org/abs/2403.05530} {Gemini 1.5: Unlocking
  multimodal understanding across millions of tokens of context}.
\newblock \emph{Preprint}, arXiv:2403.05530.

\bibitem[{Goldman et~al.(2024)Goldman, Jacovi, Slobodkin, Maimon, Dagan, and
  Tsarfaty}]{goldman2024isitreally}
Omer Goldman, Alon Jacovi, Aviv Slobodkin, Aviya Maimon, Ido Dagan, and Reut
  Tsarfaty. 2024.
\newblock \href {https://doi.org/10.18653/v1/2024.emnlp-main.924} {Is it really
  long context if all you need is retrieval? towards genuinely difficult long
  context {NLP}}.
\newblock In \emph{Proceedings of the 2024 Conference on Empirical Methods in
  Natural Language Processing}, pages 16576--16586, Miami, Florida, USA.
  Association for Computational Linguistics.

\bibitem[{Hsieh et~al.(2024)Hsieh, Sun, Kriman, Acharya, Rekesh, Jia, Zhang,
  and Ginsburg}]{hsieh2024ruler}
Cheng-Ping Hsieh, Simeng Sun, Samuel Kriman, Shantanu Acharya, Dima Rekesh, Fei
  Jia, Yang Zhang, and Boris Ginsburg. 2024.
\newblock \href {https://arxiv.org/abs/2404.06654} {Ruler: What's the real
  context size of your long-context language models?}
\newblock \emph{arXiv preprint arXiv:2404.06654}.

\bibitem[{Huang et~al.(2025)Huang, Ling, Zhong, Wu, and
  Lin}]{huang2025minilongbench}
Zhongzhan Huang, Guoming Ling, Shanshan Zhong, Hefeng Wu, and Liang Lin. 2025.
\newblock \href {https://doi.org/10.18653/v1/2025.acl-long.560}
  {{M}ini{L}ong{B}ench: The low-cost long context understanding benchmark for
  large language models}.
\newblock In \emph{Proceedings of the 63rd Annual Meeting of the Association
  for Computational Linguistics (Volume 1: Long Papers)}, pages 11442--11460,
  Vienna, Austria. Association for Computational Linguistics.

\bibitem[{Jacovi et~al.(2023)Jacovi, Caciularu, Goldman, and
  Goldberg}]{jacovi2023contamination}
Alon Jacovi, Avi Caciularu, Omer Goldman, and Yoav Goldberg. 2023.
\newblock \href {https://doi.org/10.18653/v1/2023.emnlp-main.308} {Stop
  uploading test data in plain text: Practical strategies for mitigating data
  contamination by evaluation benchmarks}.
\newblock In \emph{Proceedings of the 2023 Conference on Empirical Methods in
  Natural Language Processing}, pages 5075--5084.

\bibitem[{Jiayang et~al.(2026)Jiayang, Ru, Qiu, Li, Cao, Song, and
  Cai}]{jiayang2026amemgyminteractivememorybenchmarking}
Cheng Jiayang, Dongyu Ru, Lin Qiu, Yiyang Li, Xuezhi Cao, Yangqiu Song, and
  Xunliang Cai. 2026.
\newblock \href {https://openreview.net/forum?id=sfrVLzsmlf} {{AM}emgym:
  Interactive memory benchmarking for assistants in long-horizon
  conversations}.
\newblock In \emph{The Fourteenth International Conference on Learning
  Representations}.

\bibitem[{Kamradt(2023)}]{kamradt2023niah}
Greg Kamradt. 2023.
\newblock Needle in a haystack -- pressure testing llms.
\newblock \url{https://github.com/gkamradt/LLMTest_NeedleInAHaystack}.

\bibitem[{Kuratov et~al.(2024)Kuratov, Bulatov, Anokhin, Rodkin, Sorokin,
  Sorokin, and Burtsev}]{kuratov2024babilong}
Yuri Kuratov, Aydar Bulatov, Petr Anokhin, Ivan Rodkin, Dmitry~Igorevich
  Sorokin, Artyom Sorokin, and Mikhail Burtsev. 2024.
\newblock \href {https://openreview.net/forum?id=u7m2CG84BQ} {{BABILong}:
  Testing the limits of {LLMs} with long context reasoning-in-a-haystack}.
\newblock In \emph{The Thirty-eight Conference on Neural Information Processing
  Systems Datasets and Benchmarks Track}.

\bibitem[{Lee et~al.(2024)Lee, Chen, Dai, Dua, Sachan, Boratko, Luan, Arnold,
  Perot, Dalmia, Hu, Lin, Pasupat, Amini, Cole, Riedel, Naim, Chang, and
  Guu}]{Lee2024LongContext}
Jinhyuk Lee, Anthony Chen, Zhuyun Dai, Dheeru Dua, Devendra~Singh Sachan,
  Michael Boratko, Yi~Luan, Sebastien M.~R. Arnold, Vincent Perot, Siddharth
  Dalmia, Hexiang Hu, Xudong Lin, Panupong Pasupat, Aida Amini, Jeremy~R. Cole,
  Sebastian Riedel, Iftekhar Naim, Ming-Wei Chang, and Kelvin Guu. 2024.
\newblock \href {https://arxiv.org/abs/2406.13121} {Can long-context language
  models subsume retrieval, rag, sql, and more?}
\newblock \emph{arXiv preprint arXiv:2406.13121}.

\bibitem[{Levy et~al.(2024)Levy, Jacoby, and Goldberg}]{levy2024sametask}
Mosh Levy, Alon Jacoby, and Yoav Goldberg. 2024.
\newblock \href {https://doi.org/10.18653/v1/2024.acl-long.818} {Same task,
  more tokens: the impact of input length on the reasoning performance of large
  language models}.
\newblock In \emph{Proceedings of the 62nd Annual Meeting of the Association
  for Computational Linguistics (Volume 1: Long Papers)}, pages 15339--15353,
  Bangkok, Thailand. Association for Computational Linguistics.

\bibitem[{Lewis et~al.(2020)Lewis, Perez, Piktus, Petroni, Karpukhin, Goyal,
  K\"{u}ttler, Lewis, Yih, Rockt\"{a}schel, Riedel, and Kiela}]{lewis2020rag}
Patrick Lewis, Ethan Perez, Aleksandra Piktus, Fabio Petroni, Vladimir
  Karpukhin, Naman Goyal, Heinrich K\"{u}ttler, Mike Lewis, Wen-tau Yih, Tim
  Rockt\"{a}schel, Sebastian Riedel, and Douwe Kiela. 2020.
\newblock \href
  {https://proceedings.neurips.cc/paper_files/paper/2020/file/6b493230205f780e1bc26945df7481e5-Paper.pdf}
  {Retrieval-augmented generation for knowledge-intensive nlp tasks}.
\newblock In \emph{Advances in Neural Information Processing Systems},
  volume~33, pages 9459--9474. Curran Associates, Inc.

\bibitem[{Lin(2004)}]{lin2004rouge}
Chin-Yew Lin. 2004.
\newblock \href {https://aclanthology.org/W04-1013/} {{ROUGE}: A package for
  automatic evaluation of summaries}.
\newblock In \emph{Text Summarization Branches Out}, pages 74--81, Barcelona,
  Spain. Association for Computational Linguistics.

\bibitem[{Liu et~al.(2025)Liu, Zhu, Bai, He, Liao, Que, Wang, Zhang, Zhang,
  Zhang, Zhang, Chen, Guo, Li, Liu, Shan, Song, Tian, Wu, Zhou, Zhu, Feng, Gao,
  He, Li, Liu, Meng, Su, Tan, Wang, Yang, Ye, Zheng, Zhou, Huang, Li, and
  Zhang}]{liu2025comprehensivesurveylongcontext}
Jiaheng Liu, Dawei Zhu, Zhiqi Bai, Yancheng He, Huanxuan Liao, Haoran Que,
  Zekun Wang, Chenchen Zhang, Ge~Zhang, Jiebin Zhang, Yuanxing Zhang, Zhuo
  Chen, Hangyu Guo, Shilong Li, Ziqiang Liu, Yong Shan, Yifan Song, Jiayi Tian,
  Wenhao Wu, and 18 others. 2025.
\newblock \href {https://arxiv.org/abs/2503.17407} {A comprehensive survey on
  long context language modeling}.
\newblock \emph{Preprint}, arXiv:2503.17407.

\bibitem[{Liu et~al.(2024)Liu, Lin, Hewitt, Paranjape, Bevilacqua, Petroni, and
  Liang}]{liu2023lostmiddlelanguagemodels}
Nelson~F. Liu, Kevin Lin, John Hewitt, Ashwin Paranjape, Michele Bevilacqua,
  Fabio Petroni, and Percy Liang. 2024.
\newblock \href {https://doi.org/10.1162/tacl_a_00638} {Lost in the middle: How
  language models use long contexts}.
\newblock \emph{Transactions of the Association for Computational Linguistics},
  12:157--173.

\bibitem[{{OpenAI}(2023)}]{openai2023gpt4}
{OpenAI}. 2023.
\newblock \href {https://arxiv.org/abs/2303.08774} {Gpt-4 technical report}.
\newblock \emph{Preprint}, arXiv:2303.08774.

\bibitem[{{OpenAI}(2025{\natexlab{a}})}]{openai2025graphwalks}
{OpenAI}. 2025{\natexlab{a}}.
\newblock \href {https://huggingface.co/datasets/openai/graphwalks}
  {{GraphWalks}: a multi hop reasoning long context benchmark}.
\newblock Hugging Face dataset.

\bibitem[{{OpenAI}(2025{\natexlab{b}})}]{openai2025mrcr}
{OpenAI}. 2025{\natexlab{b}}.
\newblock \href {https://huggingface.co/datasets/openai/mrcr} {{OpenAI MRCR}:
  Long context multiple needle in a haystack benchmark}.
\newblock Hugging Face dataset.

\bibitem[{Rando et~al.(2025)Rando, Romani, Sampieri, Franco, Yang, Kyuragi,
  Galasso, and Hashimoto}]{rando2025longcodebench}
Stefano Rando, Luca Romani, Alessio Sampieri, Luca Franco, John Yang, Yuta
  Kyuragi, Fabio Galasso, and Tatsunori Hashimoto. 2025.
\newblock \href {https://arxiv.org/abs/2505.07897} {Longcodebench: Evaluating
  coding llms at 1m context windows}.
\newblock \emph{Preprint}, arXiv:2505.07897.

\bibitem[{Shaham et~al.(2023)Shaham, Ivgi, Efrat, Berant, and
  Levy}]{shaham2023zeroscrolls}
Uri Shaham, Maor Ivgi, Avia Efrat, Jonathan Berant, and Omer Levy. 2023.
\newblock \href {https://doi.org/10.18653/v1/2023.findings-emnlp.536}
  {Zeroscrolls: A zero-shot benchmark for long text understanding}.
\newblock In \emph{Findings of the Association for Computational Linguistics:
  EMNLP 2023}, pages 7977--7989.

\bibitem[{Shaham et~al.(2022)Shaham, Segal, Ivgi, Efrat, Yoran, Haviv, Gupta,
  Xiong, Geva, Berant, and Levy}]{shaham2022scrolls}
Uri Shaham, Elad Segal, Maor Ivgi, Avia Efrat, Ori Yoran, Adi Haviv, Ankit
  Gupta, Wenhan Xiong, Mor Geva, Jonathan Berant, and Omer Levy. 2022.
\newblock \href {https://doi.org/10.18653/v1/2022.emnlp-main.823} {Scrolls:
  Standardized comparison over long language sequences}.
\newblock In \emph{Proceedings of the 2022 Conference on Empirical Methods in
  Natural Language Processing}, pages 12007--12021.

\bibitem[{Wei et~al.(2023)Wei, Wang, Schuurmans, Bosma, Ichter, Xia, Chi, Le,
  and Zhou}]{wei2022chainofthought}
Jason Wei, Xuezhi Wang, Dale Schuurmans, Maarten Bosma, Brian Ichter, Fei Xia,
  Ed~Chi, Quoc Le, and Denny Zhou. 2023.
\newblock \href {https://arxiv.org/abs/2201.11903} {Chain-of-thought prompting
  elicits reasoning in large language models}.
\newblock \emph{Preprint}, arXiv:2201.11903.

\bibitem[{Yen et~al.(2025)Yen, Gao, Hou, Ding, Fleischer, Izsak, Wasserblat,
  and Chen}]{yen2025helmet}
Howard Yen, Tianyu Gao, Minmin Hou, Ke~Ding, Daniel Fleischer, Peter Izsak,
  Moshe Wasserblat, and Danqi Chen. 2025.
\newblock \href {https://openreview.net/forum?id=293V3bJbmE} {{HELMET}: How to
  evaluate long-context models effectively and thoroughly}.
\newblock In \emph{The Thirteenth International Conference on Learning
  Representations}.

\bibitem[{Zhang et~al.(2024)Zhang, Chen, Hu, Xu, Chen, Hao, Han, Thai, Wang,
  Liu, and Sun}]{zhang2024inftybench}
Xinrong Zhang, Yingfa Chen, Shengding Hu, Zihang Xu, Junhao Chen, Moo Hao,
  Xu~Han, Zhen Thai, Shuo Wang, Zhiyuan Liu, and Maosong Sun. 2024.
\newblock \href {https://doi.org/10.18653/v1/2024.acl-long.814}
  {$\infty${B}ench: Extending long context evaluation beyond 100{K} tokens}.
\newblock In \emph{Proceedings of the 62nd Annual Meeting of the Association
  for Computational Linguistics (Volume 1: Long Papers)}, pages 15262--15277,
  Bangkok, Thailand. Association for Computational Linguistics.

\bibitem[{Zheng et~al.(2023)Zheng, Chiang, Sheng, Zhuang, Wu, Zhuang, Lin, Li,
  Li, Xing, Zhang, Gonzalez, and Stoica}]{zheng2024judging}
Lianmin Zheng, Wei-Lin Chiang, Ying Sheng, Siyuan Zhuang, Zhanghao Wu, Yonghao
  Zhuang, Zi~Lin, Zhuohan Li, Dacheng Li, Eric Xing, Hao Zhang, Joseph~E.
  Gonzalez, and Ion Stoica. 2023.
\newblock \href {https://openreview.net/forum?id=uccHPGDlao} {Judging
  llm-as-a-judge with mt-bench and chatbot arena}.
\newblock In \emph{Thirty-seventh Conference on Neural Information Processing
  Systems Datasets and Benchmarks Track}.

\bibitem[{Zhou et~al.(2025)Zhou, Liu, Chen, Tian, and
  Chen}]{zhou2025gsminfinite}
Yang Zhou, Hongyi Liu, Zhuoming Chen, Yuandong Tian, and Beidi Chen. 2025.
\newblock \href {https://arxiv.org/abs/2502.05252} {Gsm-infinite: How do your
  llms behave over infinitely increasing context length and reasoning
  complexity?}
\newblock \emph{Preprint}, arXiv:2502.05252.

\end{thebibliography}

\newpage

\appendix

The appendix is organized in three parts. Appendices~\ref{app:component_details}--\ref{app:component_validation} describe the benchmark components and validate component selection. Appendices~\ref{app:weights}--\ref{app:ci} detail the scoring methodology and robustness checks. Appendices~\ref{app:leaderboard}--\ref{app:radar} present the full leaderboards, screening analyses, and extended empirical analyses.

\section{Component Descriptions}
\label{app:component_details}

This appendix provides detailed descriptions of each benchmark component in \atlas, including construction methodology, length extension procedures where applicable, and selection rationale.

\paragraph{MRCR-8 needle~\citep{openai2025mrcr}.}
Multi-Round Co-Reference Resolution (MRCR) is a synthetic retrieval benchmark released by OpenAI. Each instance embeds a chain of co-referential needles at controlled positions throughout a long context and asks the model to resolve the chain end-to-end. The original dataset supports 2-, 4-, and 8-needle variants; \atlas retains only the 8-needle setting, which maximizes difficulty and cross-model discrimination. Scoring uses exact match (EM) on the resolved entity. Because multiple instances share the same base context, confidence intervals are computed with the cluster-based correction described in Appendix~\ref{app:ci}.

\paragraph{OOLong-Synth~\citep{bertsch2025oolongevaluatinglongcontext}.}
OOLong-Synth (OOLong, synthetic split) evaluates scattered-information extraction: the model must locate and aggregate multiple related facts distributed across distant segments of the input. It replaces RULER's aggregation tasks, which we found to be saturated for frontier models. Scoring uses an answer-level score: categorical and date answers require exact matches after lightweight parsing, numeric answers receive exponentially decayed partial credit based on absolute error, and frequency-comparison answers are matched against the normalized label. Instances are approximately independent, so standard CLT-based confidence intervals apply.

\paragraph{GraphWalks Extend~\citep{openai2025graphwalks}.}
GraphWalks is a synthetic graph reasoning benchmark released by OpenAI. Each instance presents a directed graph as an edge list and asks the model to perform an operation---either breadth-first search (BFS) at a specified depth or parent-node enumeration. The graph size controls prompt length. The official release provides data at only 128K and 1M context lengths; we synthesized the remaining six \atlas slices (8K, 16K, 32K, 64K, 256K, 512K) using the same generation procedure described in the official dataset documentation. Scoring uses F1 between the predicted and gold node sets.

\paragraph{LOFT-Text Retrieval Extend~\citep{Lee2024LongContext,yen2025helmet}.}
LOFT-Text Retrieval is a retrieval-oriented QA component that combines retrieval scenarios from LOFT~\citep{Lee2024LongContext} with RAG question-answering instances from HELMET~\citep{yen2025helmet}. LOFT embeds gold passages into a large corpus of distractor passages and asks the model to identify the relevant evidence for a query; HELMET-RAG follows a similar evidence-grounding paradigm with its own question sets. The official LOFT release covers 32K, 128K, and 1M context lengths; we kept these slices intact and obtained the remaining five slices (8K, 16K, 64K, 256K, 512K) by downsampling passages from the 1M Retrieval subset. Scoring uses LOFT's MRecall@K: an instance receives 1 only when the top-$K$ predicted IDs contain $\min(K,|\mathcal{G}|)$ distinct gold IDs from the gold set $\mathcal{G}$, and 0 otherwise.

\paragraph{Helmet-ICL Extend~\citep{yen2025helmet}.}
Helmet-ICL evaluates in-context learning by presenting a large block of labeled demonstrations followed by a test query. It was selected from multiple candidate ICL tasks because it exhibits high cross-model discrimination (standard deviation $\sigma=24.0$ at 128K and $\sigma=31.8$ at 1M across 26 models). Scoring uses classification accuracy (Acc). The official HELMET release provides construction scripts that generate ICL instances at configurable context lengths; we used these scripts\footnote{\url{https://github.com/princeton-nlp/HELMET}} to produce instances at the \atlas length slices not covered by the original release.

\paragraph{LongCodeBench~\citep{rando2025longcodebench}.}
LongCodeBench tests repository-scale code understanding through two subtasks: LongCodeQA (natural-language questions about repository logic) and LongSWE (software engineering modifications). Because repository-scale inputs have a natural minimum length, coverage begins at 32K rather than 8K, and instance counts vary across slices (see Table~\ref{tab:subset_stats}). Scoring follows each subtask's native verifier: LongCodeQA is scored by exact matching the predicted option letter against the gold answer, while LongSWE is scored by functional resolution of a single generated patch (Pass@1). The reported LongCodeBench score is the instance-count-weighted mean of binary success scores across the two subtasks. LongCodeBench has comparatively low cross-model separation ($\sigma=10.9$ at 128K) and the lowest correlations with other dimensions, indicating that it captures a relatively independent capability.

\paragraph{AMemBench-ACU~\citep{jiayang2026amemgyminteractivememorybenchmarking}.}
AMemBench-ACU measures long-range memory from ultra-long conversational streams. The original Amembench benchmark provides an interactive memory evaluation paradigm; we extended it to the \atlas length slices by having frontier models generate multi-turn interaction processes at the required context lengths, then converting the resulting transcripts into static evaluation instances. Scoring uses quasi-prefix exact match (QPEM): after lowercasing, punctuation removal, and whitespace normalization, an instance receives 1 if the normalized prediction starts with any normalized gold ACU and 0 otherwise. For fixed output mappings, we use exact match after mapping to avoid prefix collisions between similar options. A limitation of this extension procedure is that the evaluation data is generated by language models, which introduces a potential distribution-match bias: models sharing architectural or training similarities with the generator may find the resulting conversational patterns easier to process. We partially mitigate this by using multiple generator models and verifying that the generated transcripts satisfy target length and content-diversity constraints, but the bias cannot be fully eliminated. Future releases should explore human-authored or hybrid generation approaches for the memory dimension.

\paragraph{LongBench-v2~\citep{bai2024longbenchv2} and AA-LCR~\citep{artificialanalysis2025lcr}.}
These two components form the Holistic Assessment scoring category, contributing to ATLAScore as a third harmonic-mean term alongside the Foundational and length-sliced Application aggregates. LongBench-v2 provides 503 challenging four-choice questions spanning six realistic long-context task categories: single-document QA, multi-document QA, long in-context learning, long-dialogue history understanding, code repository understanding, and long structured data understanding. Its contexts range from 8K to 2M words, and its official difficulty split contains 192 easy and 311 hard instances. AA-LCR provides cross-document analytical reasoning questions (100 instances). Both are evaluated at their original instance lengths rather than the \atlas length slices, and both use classification accuracy (Acc) as the metric. Because these component scores are not recomputed by reporting scope, their aggregate score $S(L^\star)$ is identical for both the @128K and @1M scopes.

\paragraph{Per-subset instance counts.}
Table~\ref{tab:subset_stats} provides the per-subset instance counts at each length slice, complementing the aggregate totals in Table~\ref{tab:components}. Most subsets maintain 100 instances per length slice by design. Notable exceptions include MRCR, whose instance counts vary slightly across lengths due to the constraint that multi-needle chains must fit within the target context window, and LongCodeBench, whose repository-scale inputs lead to non-uniform counts. LongBench-v2 and AA-LCR are evaluated at their original instance lengths rather than the \atlas length slices.

\begin{table*}[tp]
\caption{Per-subset instance counts at each length slice. Dashes indicate that the subset does not cover that length; LongCodeBench's unsupported 8K and 16K slices are omitted from per-slice application aggregation. Holistic assessment components (LongBench-v2, AA-LCR) are evaluated at their original instance lengths; counts are totals rather than per-slice.}
\label{tab:subset_stats}
\centering
\footnotesize
\renewcommand{\arraystretch}{1.25}
\resizebox{\textwidth}{!}{
\begin{tabular}{lccccccccr}
\toprule
Subset & Metric & 8K & 16K & 32K & 64K & 128K & 256K & 512K & 1M \\
\midrule
MRCR-8 needle & EM & 106 & 96 & 98 & 100 & 100 & 100 & 100 & 92 \\
OOLong-Synth & Answer-level & 100 & 100 & 100 & 100 & 100 & 100 & 100 & 100 \\
GraphWalks Extend & F1 & 100 & 100 & 100 & 100 & 100 & 100 & 100 & 100 \\
LOFT-Text Retrieval Extend & MRecall@K & 100 & 100 & 100 & 100 & 100 & 100 & 100 & 100 \\
Helmet-ICL Extend & Acc & 100 & 100 & 100 & 100 & 100 & 100 & 100 & 100 \\
LongCodeBench & Acc / Pass@1 & -- & -- & 213 & 176 & 192 & 165 & 147 & 150 \\
AMemBench-ACU & QPEM & 100 & 100 & 100 & 100 & 100 & 100 & 100 & 100 \\
\midrule
LongBench-v2 & Acc & \multicolumn{8}{c}{503} \\
AA-LCR & Acc & \multicolumn{8}{c}{100} \\
\bottomrule
\end{tabular}
}
\end{table*}

\section{Component Selection Validation}
\label{app:component_validation}

This appendix provides quantitative evidence that the seven length-sliced \atlas components satisfy the selection criteria stated in Section~\ref{sec:construction}: controllable length coverage across the \atlas grid where supported, sufficient cross-model discrimination, and low inter-component redundancy. LongCodeBench starts at 32K because repository-scale inputs cannot be meaningfully compressed to shorter contexts, as described in Appendix~\ref{app:component_details}. The two holistic assessment components (LongBench-v2, AA-LCR) are evaluated at original instance lengths and form a separate scoring category. We further validate the robustness of the overall ranking through a leave-one-dimension-out analysis.

\subsection{Inter-Dimension Correlation}

Table~\ref{tab:corr_matrix} reports the pairwise Spearman rank correlations among the seven length-sliced capability dimensions at 128K across all 26 models. The mean off-diagonal correlation is $\bar{\rho} = 0.64$ (median $0.76$), with 6 of 21 unrounded pairs falling below $\rho = 0.50$ and none exceeding $\rho = 0.90$. Code is the most independent dimension, with a maximum pairwise $\rho$ of about $0.49$; its correlations with Retrieval ($\rho = 0.07$, $p = 0.746$) and Reasoning ($\rho = 0.17$, $p = 0.416$) are not statistically significant, confirming that code understanding captures a distinct capability not reducible to retrieval or reasoning strength. Correlations within the foundational layer (mean $\rho = 0.82$) are higher than within the application layer (mean $\rho = 0.59$) or across layers (mean $\rho = 0.63$), consistent with the two-layer design in which foundational dimensions share a common synthetic, knowledge-free structure while application dimensions target heterogeneous downstream capabilities. At the 1M scope, the overall correlation structure is similar (mean $\bar{\rho} = 0.68$, median $0.76$), with five dimension pairs below $\rho = 0.50$.

\begin{table*}[tp]
\caption{Pairwise Spearman $\rho$ among the seven length-sliced capability dimensions at 128K ($N=26$ models). Values are rounded to two decimals; threshold counts in the text use unrounded values. Significance: $^{**}$\,$p < 0.01$, $^{*}$\,$p < 0.05$; unmarked pairs are not significant at $p = 0.05$. Dimensions above the mid-rule are foundational; below are application.}
\label{tab:corr_matrix}
\centering
\footnotesize
\renewcommand{\arraystretch}{1.25}
\setlength{\tabcolsep}{4.5pt}
\resizebox{\textwidth}{!}{
\begin{tabular}{lccccccc}
\toprule
 & Retrieval & Aggregation & Reasoning & QA & ICL & Code & Memory \\
\midrule
Retrieval          & 1.00 & 0.80$^{**}$ & 0.86$^{**}$ & 0.70$^{**}$ & 0.80$^{**}$ & 0.07        & 0.69$^{**}$ \\
Aggregation       &      & 1.00       & 0.79$^{**}$ & 0.87$^{**}$ & 0.82$^{**}$ & 0.39$^{*}$  & 0.87$^{**}$ \\
Reasoning        &      &            & 1.00       & 0.71$^{**}$ & 0.76$^{**}$ & 0.17        & 0.72$^{**}$ \\
\midrule
QA       &      &            &            & 1.00       & 0.77$^{**}$ & 0.45$^{*}$  & 0.80$^{**}$ \\
ICL         &      &            &            &            & 1.00       & 0.25        & 0.77$^{**}$ \\
Code       &      &            &            &            &            & 1.00       & 0.49$^{*}$ \\
Memory          &      &            &            &            &            &            & 1.00       \\
\bottomrule
\end{tabular}
}
\end{table*}

\begin{figure*}[tp]
\centering
\begin{subfigure}[b]{0.48\textwidth}
    \centering
    \includegraphics[width=\linewidth]{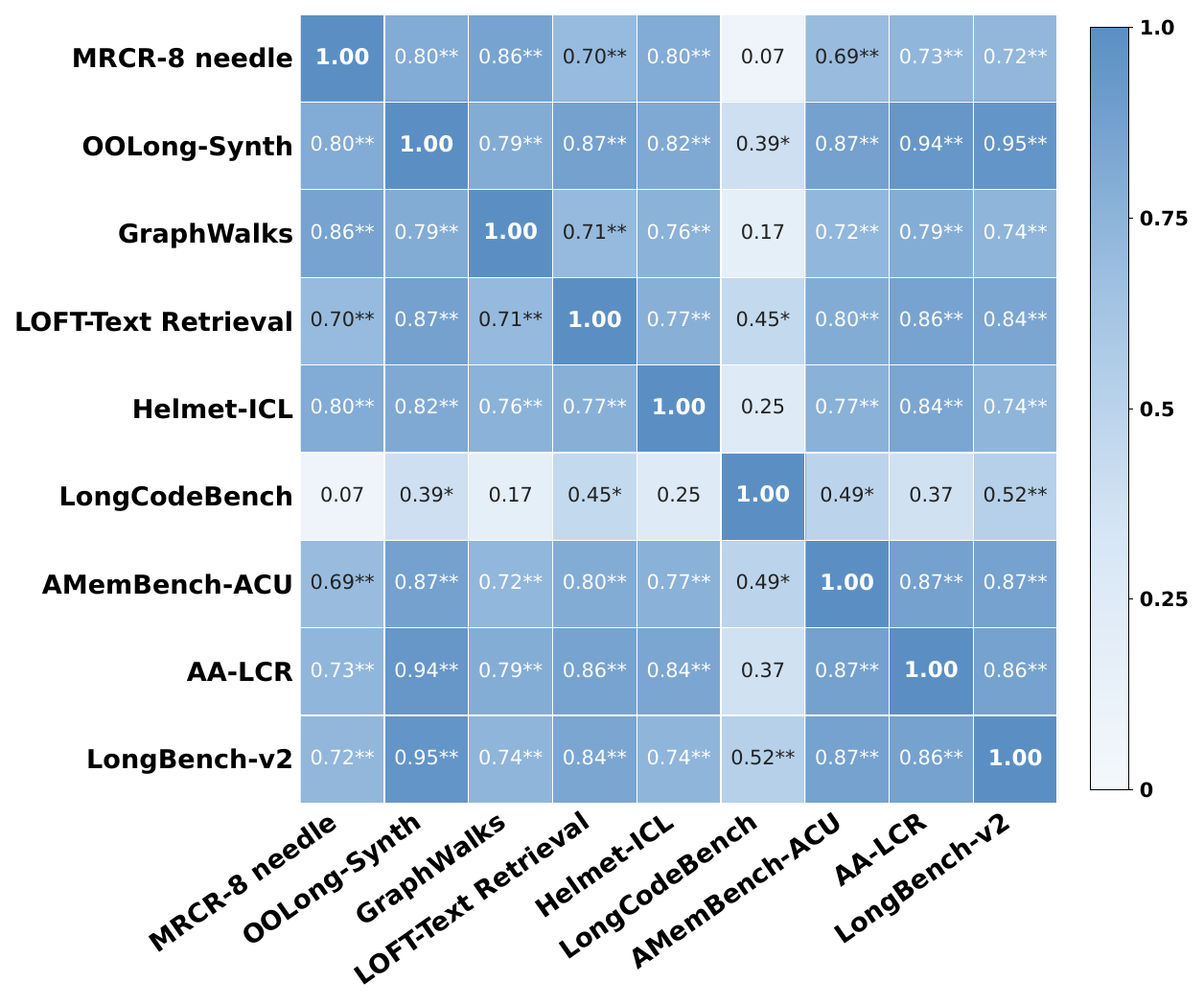}
    \caption{128K Context}
\end{subfigure}
\hfill
\begin{subfigure}[b]{0.48\textwidth}
    \centering
    \includegraphics[width=\linewidth]{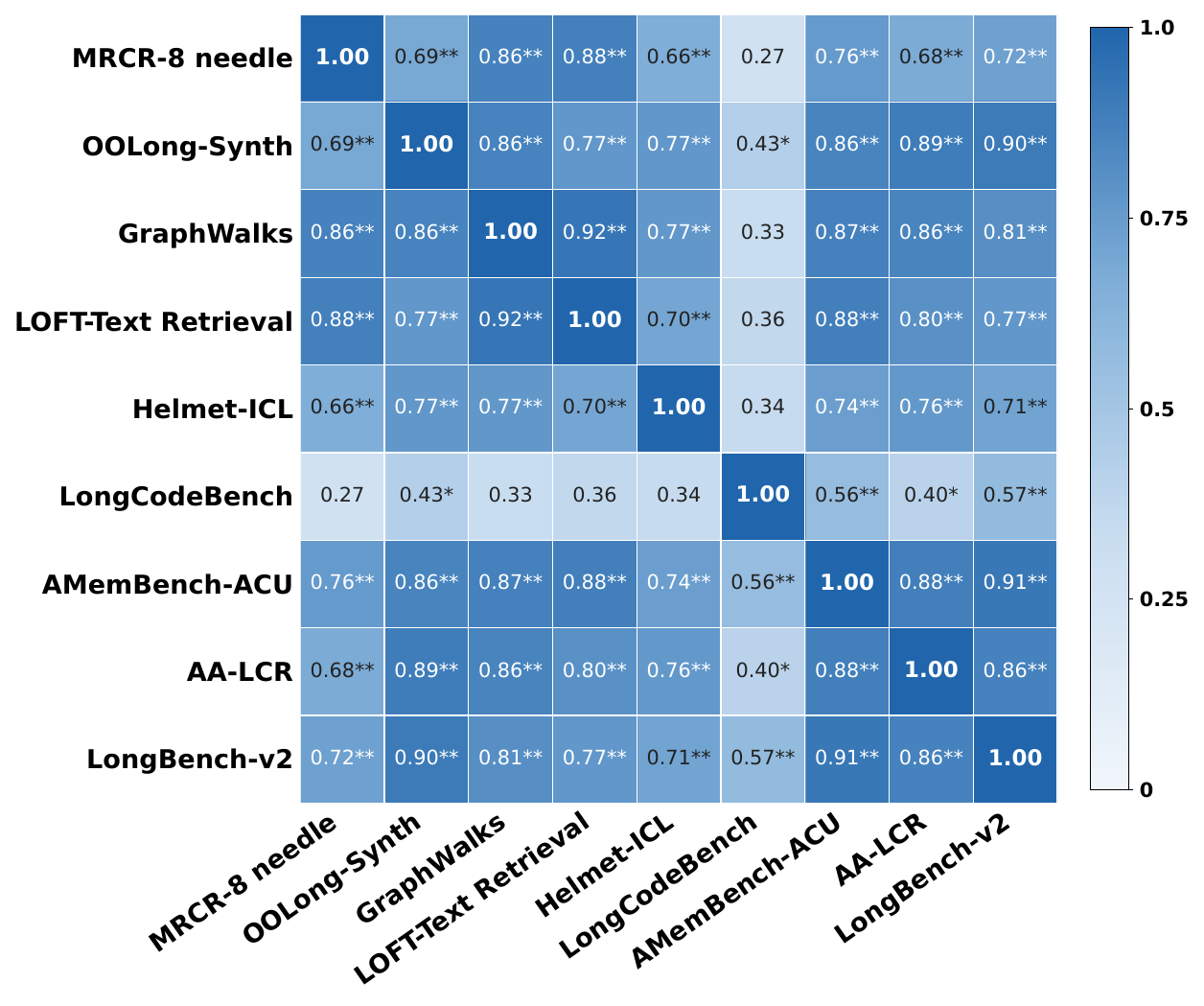}
    \caption{1M Context}
\end{subfigure}
\caption{Pairwise Spearman $\rho$ among all nine \atlas components (seven length-sliced + two holistic assessment) at 128K~(a) and 1M~(b). Significance levels are annotated in each cell (* $p<0.05$, ** $p<0.01$).}
\label{fig:corr_heatmap}
\end{figure*}

\begin{figure*}[tp]
\centering
\begin{subfigure}[b]{0.48\textwidth}
    \centering
    \includegraphics[width=\linewidth]{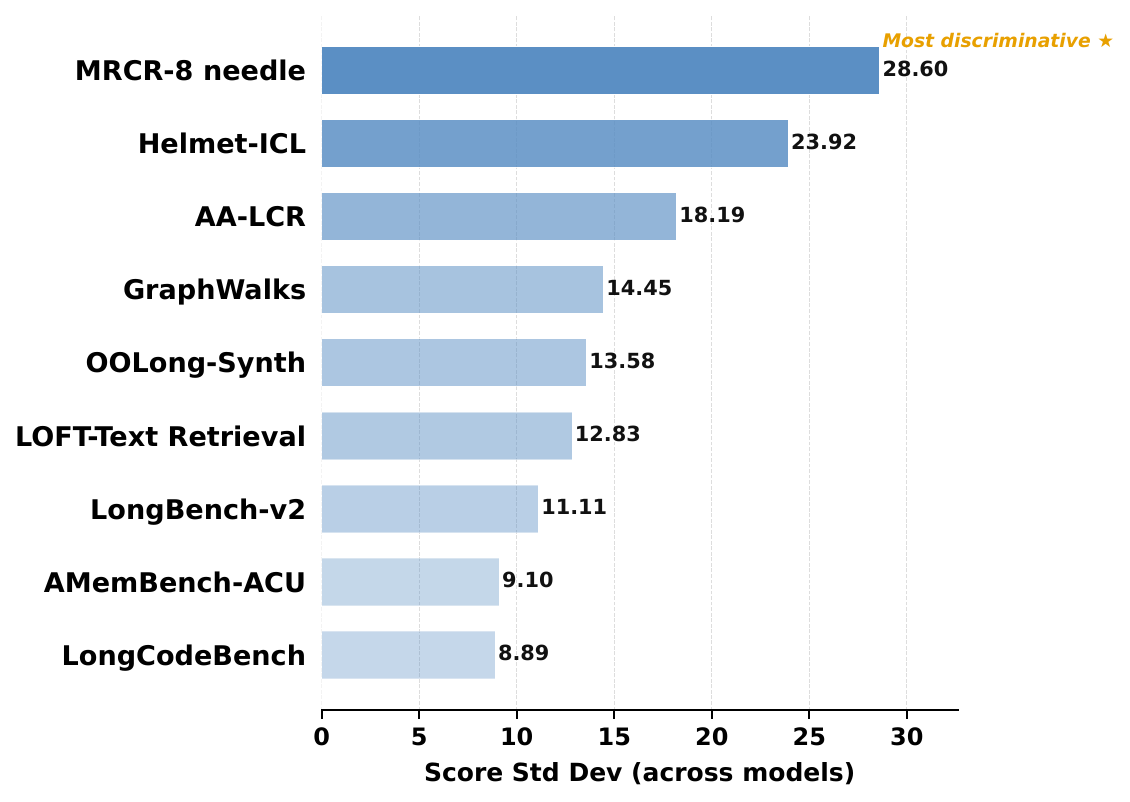}
    \caption{128K Context}
\end{subfigure}
\hfill
\begin{subfigure}[b]{0.48\textwidth}
    \centering
    \includegraphics[width=\linewidth]{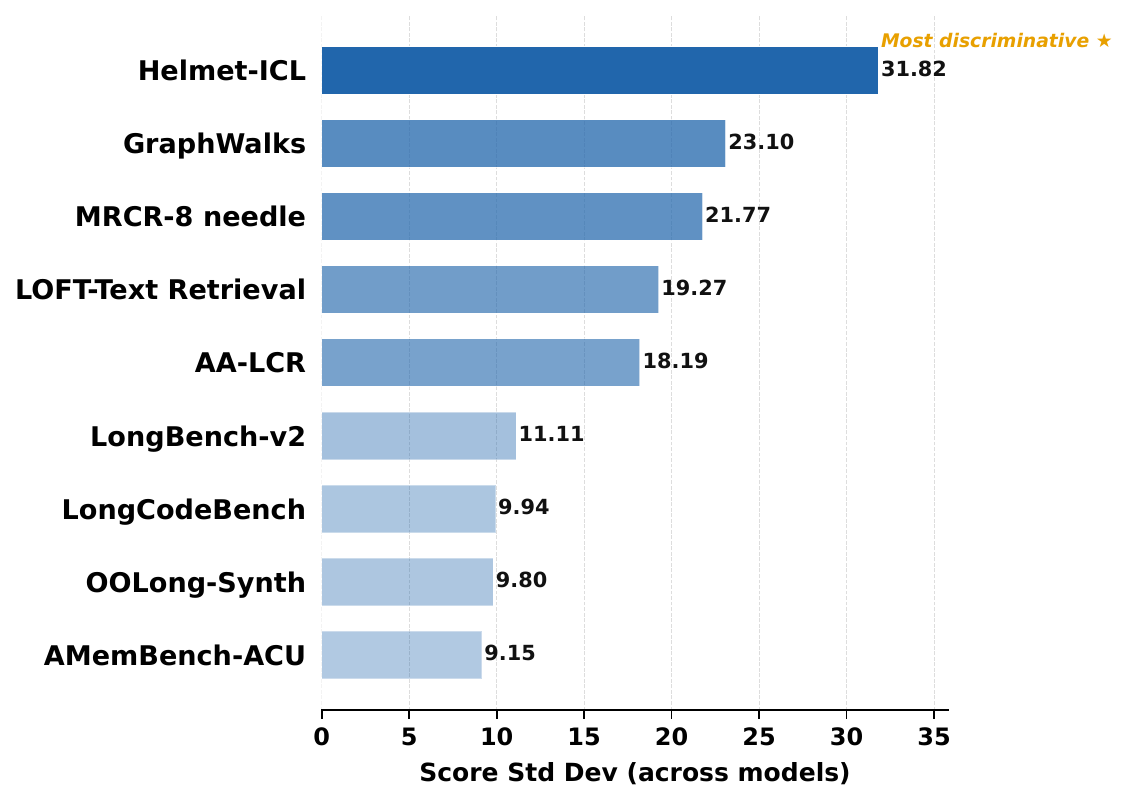}
    \caption{1M Context}
\end{subfigure}
\caption{Per-component discriminability (cross-model score standard deviation) at 128K~(a) and 1M~(b), including both length-sliced and holistic assessment components.}
\label{fig:discriminability}
\end{figure*}

Figure~\ref{fig:corr_heatmap} visualizes the full correlation matrices at both 128K and 1M for all nine components (seven length-sliced plus two holistic assessment), and Figure~\ref{fig:discriminability} shows the per-component discriminability.

\subsection{Cross-Model Discriminability}

Table~\ref{tab:discriminability} reports the cross-model standard deviation ($\sigma$) for each dimension, measuring how well the dimension separates strong from weak long-context models. All seven length-sliced dimensions achieve $\sigma \geq 8.9$ at 128K and $\sigma \geq 9.2$ at 1M, confirming that every dimension contributes meaningful discriminative power to the benchmark. Retrieval ($\sigma = 28.6$) and ICL ($\sigma = 23.9$) are the two most discriminative dimensions at 128K, while ICL ($\sigma = 31.8$) and Reasoning ($\sigma = 23.1$) lead at 1M. Even the least discriminative dimension, Code ($\sigma = 8.9$ at 128K), retains sufficient spread for meaningful model differentiation.

\begin{table}[t]
\caption{Cross-model discriminability ($\sigma$, standard deviation of scores across 26 models) for each length-sliced capability dimension, sorted by $\sigma$ at 128K.}
\label{tab:discriminability}
\centering
\small
\renewcommand{\arraystretch}{1.2}
\setlength{\tabcolsep}{3pt}
\resizebox{\columnwidth}{!}{
\begin{tabular}{llcc}
\toprule
Dimension & Layer & $\sigma$ @128K & $\sigma$ @1M \\
\midrule
Retrieval          & Foundational & 28.60 & 21.77 \\
ICL    & Application  & 23.92 & 31.82 \\
Reasoning    & Foundational & 14.45 & 23.10 \\
Aggregation & Foundational & 13.58 &  9.80 \\
QA & Application  & 12.83 & 19.27 \\
Memory & Application  &  9.10 &  9.15 \\
Code & Application  &  8.89 &  9.94 \\
\midrule
\textbf{Mean} &              & \textbf{15.91} & \textbf{17.83} \\
\bottomrule
\end{tabular}
}
\end{table}

\subsection{Leave-One-Dimension-Out Robustness}

To verify that no single dimension dominates or distorts the overall ranking, we perform a leave-one-dimension-out analysis: for each of the seven length-sliced dimensions, we remove it, recompute the ATLAScore using the remaining dimensions (preserving the three-category harmonic-mean aggregation), and measure the rank correlation between the reduced ranking and the full ranking across all 26 models.

Table~\ref{tab:leave_one_out} reports the results. All seven leave-one-out rankings maintain Spearman $\rho \geq 0.97$ (128K) and $\rho \geq 0.96$ (1M) with the full ranking. The mean $\rho$ across all seven ablations is $0.987$ (128K) and $0.992$ (1M), confirming that the overall ranking is robust and not dominated by any single dimension.

\begin{table*}[tp]
\caption{Leave-one-dimension-out analysis. $\rho$: Spearman rank correlation between the reduced and full ATLAScore rankings. $\tau$: Kendall rank correlation. Max~$|\Delta r|$: largest absolute rank shift for any single model under the reduced ranking.}
\label{tab:leave_one_out}
\centering
\small
\renewcommand{\arraystretch}{1.25}
\resizebox{\textwidth}{!}{
\begin{tabular}{llcccccc}
\toprule
 & & \multicolumn{3}{c}{@128K} & \multicolumn{3}{c}{@1M} \\
\cmidrule(lr){3-5} \cmidrule(lr){6-8}
Dropped Dimension & Layer & $\rho$ & $\tau$ & Max $|\Delta r|$ & $\rho$ & $\tau$ & Max $|\Delta r|$ \\
\midrule
Retrieval & Foundational & 0.977 & 0.926 & 6 & 0.992 & 0.945 & 2 \\
Aggregation & Foundational & 0.986 & 0.920 & 3 & 0.992 & 0.957 & 3 \\
Reasoning    & Foundational & 0.985 & 0.926 & 4 & 0.997 & 0.975 & 1 \\
QA & Application   & 0.990 & 0.938 & 2 & 0.998 & 0.982 & 1 \\
ICL    & Application   & 0.982 & 0.926 & 4 & 0.966 & 0.877 & 5 \\
Code & Application   & 0.991 & 0.945 & 3 & 0.997 & 0.975 & 2 \\
Memory & Application   & 0.998 & 0.988 & 2 & 0.999 & 0.994 & 1 \\
\midrule
\textbf{Mean} &        & \textbf{0.987} & \textbf{0.938} & --- & \textbf{0.992} & \textbf{0.959} & --- \\
\bottomrule
\end{tabular}
}
\end{table*}

Retrieval produces the largest perturbation at 128K when removed (maximum rank shift of 6 positions), while ICL produces the largest perturbation at 1M ($\rho = 0.966$, maximum rank shift of 5 positions). This is expected given their high discriminability and their imperfect correlation with several other dimensions. Rather than indicating benchmark fragility, this sensitivity confirms that retrieval and in-context learning capture non-redundant capability dimensions whose inclusion is empirically justified.

At the other extreme, removing Memory yields $\rho = 0.998$ at 128K and $\rho = 0.999$ at 1M, indicating lower marginal impact on relative rankings. However, each dimension remains essential for diagnostic completeness: Code is the most orthogonal dimension, Memory is the sole long-range memory probe, and Aggregation is the only aggregation-specific dimension that has not saturated for frontier models.

The top-ranked models (Gemini-3.1-Pro-Preview, Claude-Opus-4.6, and GPT-5.5) exhibit small rank shifts across the ablations, indicating that the leading positions are robust to any single dimension choice.

\section{Length-Slice Weights}
\label{app:weights}

Table~\ref{tab:weights} lists the default length-slice weights used in all experiments reported in this paper. Under uniform weights ($w_\ell = 1$), the trapezoidal AUC formula naturally assigns greater influence to wider intervals on the token-count axis, so longer-range segments (e.g., 512K$\to$1M spans 524K tokens vs.\ 8K$\to$16K spanning only 8K tokens) contribute proportionally more to the score. This is by design: the geometric grid ensures that performance at scale dominates the aggregate without requiring explicit weight tuning.

\begin{table}[t]
\caption{Default length-slice weights $w_{\ell}$ used in the \atlas AUC computation.}
\label{tab:weights}
\centering
\small
\renewcommand{\arraystretch}{1.2}
\setlength{\tabcolsep}{3pt}
\resizebox{\columnwidth}{!}{
\begin{tabular}{lcccccccc}
\toprule
Length $\ell$ & 8K & 16K & 32K & 64K & 128K & 256K & 512K & 1M \\
\midrule
$w_{\ell}$ & 1 & 1 & 1 & 1 & 1 & 1 & 1 & 1 \\
\bottomrule
\end{tabular}
}
\end{table}

Practitioners who wish to further penalize late-stage degradation may adopt a weighted variant of the trapezoidal rule by multiplying each segment by a nonnegative weight factor (e.g., proportional to $\log_2 \ell$). The scoring framework in Section~\ref{sec:method} generalizes naturally: weights modulate segment contributions without changing the trapezoidal structure.

\paragraph{Sensitivity analysis.}
A natural concern is whether the uniform-weight default is a consequential design choice: would alternative weighting schemes change the model rankings? We compare three schemes---\textbf{Uniform} ($w_\ell=1$), \textbf{Logarithmic} ($w_\ell = \log_2(\ell/\text{4K})$, which up-weights longer slices on a log scale), and \textbf{Inverse Logarithmic} ($w_\ell = \log_2(\text{2M}/\ell)$, a deployment-frequency proxy that up-weights shorter slices)---across all 26 models. Table~\ref{tab:weight_sensitivity} reports the resulting \texttt{ATLAScore} and rank under each scheme for both the @128K and @1M scopes.

Rankings are highly stable at 128K: Spearman $\rho$ between Uniform and Log is 1.000, and between Uniform and Inverse Logarithmic is 1.000. At 1M, both alternatives remain close to the default (Log: $\rho=0.999$; Inverse Logarithmic: $\rho=0.997$), with a maximum rank shift of 1. These results support the uniform default: alternative length weights can express different deployment preferences, but they do not overturn the main ordering.

\begin{table*}[tp]
\caption{Sensitivity of \texttt{ATLAScore} rankings to length-slice weighting schemes. Superscripts denote rank. Spearman $\rho$ is computed against the Uniform baseline.}
\label{tab:weight_sensitivity}
\centering
\scriptsize
\setlength{\tabcolsep}{3pt}
\resizebox{\textwidth}{!}{
\begin{tabular}{l ccc ccc}
\toprule
& \multicolumn{3}{c}{\texttt{ATLAScore@8K-128K}} & \multicolumn{3}{c}{\texttt{ATLAScore@8K-1M}} \\
\cmidrule(lr){2-4} \cmidrule(lr){5-7}
Model & Uniform & Logarithmic & Inverse Logarithmic & Uniform & Logarithmic & Inverse Logarithmic \\
\midrule
Gemini-3.1-Pro-Preview (high) & 77.8$^{1}$ & 77.2$^{1}$ & 78.3$^{1}$ & 68.5$^{2}$ & 67.5$^{2}$ & 71.0$^{2}$ \\
Claude-Opus-4.6 (max) & 77.1$^{2}$ & 76.7$^{2}$ & 77.4$^{2}$ & 70.5$^{1}$ & 69.9$^{1}$ & 72.2$^{1}$ \\
GPT-5.5 (xhigh) & 74.6$^{3}$ & 74.1$^{3}$ & 75.0$^{3}$ & 67.7$^{3}$ & 67.1$^{3}$ & 69.4$^{3}$ \\
GPT-5.2 (xhigh) & 73.7$^{4}$ & 73.2$^{4}$ & 74.1$^{4}$ & 58.2$^{8}$ & 56.5$^{8}$ & 62.6$^{7}$ \\
GPT-5.4 (xhigh) & 73.4$^{5}$ & 72.9$^{5}$ & 73.9$^{5}$ & 63.2$^{4}$ & 62.2$^{5}$ & 65.9$^{4}$ \\
Gemini-3-Pro-Preview (high) & 73.0$^{6}$ & 72.2$^{6}$ & 73.5$^{6}$ & 62.6$^{6}$ & 61.6$^{6}$ & 65.3$^{5}$ \\
Gemini-3-Flash-Preview (high) & 72.0$^{7}$ & 71.2$^{7}$ & 72.5$^{7}$ & 63.1$^{5}$ & 62.4$^{4}$ & 65.1$^{6}$ \\
DeepSeek-V4-Pro (max) & 71.6$^{8}$ & 70.9$^{8}$ & 72.0$^{8}$ & 59.1$^{7}$ & 57.8$^{7}$ & 62.4$^{8}$ \\
Qwen3.5-397B-A17B (Reasoning) & 71.0$^{9}$ & 70.4$^{9}$ & 71.5$^{9}$ & 54.4$^{10}$ & 52.7$^{10}$ & 58.9$^{10}$ \\
Claude-Opus-4.5 (high) & 69.9$^{10}$ & 69.0$^{10}$ & 70.5$^{10}$ & 51.8$^{11}$ & 50.1$^{11}$ & 56.4$^{11}$ \\
Qwen3.6-Plus (Reasoning) & 68.3$^{11}$ & 67.7$^{11}$ & 68.8$^{11}$ & 55.0$^{9}$ & 53.4$^{9}$ & 59.1$^{9}$ \\
Kimi-K2.6 (Reasoning) & 64.4$^{12}$ & 63.3$^{12}$ & 65.2$^{12}$ & 50.5$^{13}$ & 49.0$^{13}$ & 54.4$^{12}$ \\
Kimi-K2.5 (Reasoning) & 64.3$^{13}$ & 63.3$^{13}$ & 65.1$^{13}$ & 47.1$^{14}$ & 45.3$^{14}$ & 51.7$^{14}$ \\
GLM-5 (Reasoning) & 64.3$^{14}$ & 63.3$^{14}$ & 65.1$^{14}$ & 44.3$^{16}$ & 42.3$^{16}$ & 49.2$^{16}$ \\
GLM-5.1 (Reasoning) & 63.9$^{15}$ & 62.9$^{15}$ & 64.6$^{15}$ & 46.2$^{15}$ & 44.5$^{15}$ & 50.6$^{15}$ \\
Qwen3.5-397B-A17B (Non-reasoning) & 63.1$^{16}$ & 62.4$^{16}$ & 63.6$^{16}$ & 50.6$^{12}$ & 49.3$^{12}$ & 54.1$^{13}$ \\
Kimi-K2.5 (Non-reasoning) & 60.1$^{17}$ & 59.1$^{17}$ & 60.7$^{17}$ & 40.6$^{17}$ & 38.4$^{17}$ & 46.0$^{17}$ \\
GLM-5 (Non-reasoning) & 49.5$^{18}$ & 48.6$^{18}$ & 50.1$^{18}$ & 37.3$^{18}$ & 36.3$^{18}$ & 40.1$^{18}$ \\
Kimi-K2-0905-Instruct & 47.6$^{19}$ & 46.8$^{19}$ & 48.2$^{19}$ & 33.9$^{20}$ & 32.4$^{20}$ & 37.7$^{20}$ \\
Qwen3-235B-A22B-2507-Instruct & 46.6$^{20}$ & 45.9$^{20}$ & 47.1$^{20}$ & 35.0$^{19}$ & 33.8$^{19}$ & 38.0$^{19}$ \\
DeepSeek-V3.1 (Non-reasoning) & 45.6$^{21}$ & 44.3$^{21}$ & 46.4$^{21}$ & 28.5$^{23}$ & 26.8$^{23}$ & 32.7$^{22}$ \\
Qwen3-Next-80B-A3B-Instruct & 44.6$^{22}$ & 43.4$^{22}$ & 45.4$^{22}$ & 28.8$^{22}$ & 27.2$^{22}$ & 32.7$^{23}$ \\
DeepSeek-V3.2 (Non-reasoning) & 43.5$^{23}$ & 42.5$^{23}$ & 44.2$^{23}$ & 27.6$^{24}$ & 26.1$^{24}$ & 31.5$^{24}$ \\
Kimi-Linear-48B-A3B-Instruct & 42.4$^{24}$ & 42.0$^{24}$ & 42.7$^{24}$ & 33.1$^{21}$ & 32.0$^{21}$ & 35.8$^{21}$ \\
GLM-4.7 (Non-reasoning) & 39.9$^{25}$ & 38.9$^{25}$ & 40.6$^{25}$ & 15.8$^{26}$ & 13.0$^{26}$ & 21.9$^{26}$ \\
Qwen3-30B-A3B-2507-Instruct & 33.4$^{26}$ & 32.3$^{26}$ & 34.1$^{26}$ & 24.5$^{25}$ & 23.7$^{25}$ & 26.6$^{25}$ \\
\midrule
Spearman $\rho$ (Uniform\,vs\,Logarithmic) & \multicolumn{3}{c}{$\rho = 1.000$} & \multicolumn{3}{c}{$\rho = 0.999$} \\
Spearman $\rho$ (Uniform\,vs\,Inverse Logarithmic) & \multicolumn{3}{c}{$\rho = 1.000$} & \multicolumn{3}{c}{$\rho = 0.997$} \\
\bottomrule
\end{tabular}
}
\end{table*}

\section{Aggregation Method Sensitivity}
\label{app:aggregation}

\atlas aggregates the three scoring-category scores ($B$, $C$, $S$) using a harmonic mean by default. A natural question is whether this choice materially affects model rankings relative to simpler alternatives. We compare four aggregation functions applied to the same category scores across all 26 models: \textbf{Harmonic mean} ($3/(1/B + 1/C + 1/S)$, the default), \textbf{Arithmetic mean} ($(B+C+S)/3$), \textbf{Geometric mean} ($(BCS)^{1/3}$), and \textbf{Minimum} ($\min(B,C,S)$).

Table~\ref{tab:aggregation_sensitivity} reports the ranking stability of each alternative relative to the harmonic-mean baseline. The arithmetic and geometric means produce near-identical rankings (Spearman $\rho \geq 0.995$ at both scopes, maximum rank shift $\leq 3$), confirming that the top-level aggregation rule is not a consequential design choice for relative model comparison. The minimum function shows slightly more divergence ($\rho = 0.945$ at 128K and $\rho = 0.988$ at 1M), because it amplifies the effect of a single weak category and therefore exposes imbalanced profiles more aggressively.

The practical difference between harmonic and arithmetic aggregation is one of diagnostic philosophy rather than ranking fidelity: the harmonic mean penalizes extreme weakness more heavily (a model scoring 90/90/30 receives 50.6 under harmonic vs.\ 70.0 under arithmetic), which aligns with the design goal of preventing inflated rankings from imbalanced profiles. Since this penalty does not materially alter relative rankings ($\rho \geq 0.995$ for arithmetic and geometric aggregation), the choice is a conservative design default rather than a critical assumption.

\begin{table*}[tp]
\caption{Sensitivity of ATLAScore rankings to the choice of aggregation method. All methods aggregate the same three category scores ($B$, $C$, $S$). Spearman $\rho$ and Kendall $\tau$ are computed against the default harmonic mean. Max~$|\Delta r|$ is the largest absolute rank shift for any single model relative to the default ranking.}
\label{tab:aggregation_sensitivity}
\centering
\small
\renewcommand{\arraystretch}{1.25}
\resizebox{\textwidth}{!}{
\begin{tabular}{lcccccc}
\toprule
& \multicolumn{3}{c}{@8K-128K} & \multicolumn{3}{c}{@8K-1M} \\
\cmidrule(lr){2-4} \cmidrule(lr){5-7}
Aggregation & $\rho$ & $\tau$ & Max $|\Delta r|$ & $\rho$ & $\tau$ & Max $|\Delta r|$ \\
\midrule
Harmonic mean (default) & 1.000 & 1.000 & 0 & 1.000 & 1.000 & 0 \\
Arithmetic mean & 0.995 & 0.975 & 3 & 0.997 & 0.975 & 1 \\
Geometric mean & 0.998 & 0.988 & 2 & 0.999 & 0.988 & 1 \\
Minimum & 0.945 & 0.809 & 7 & 0.988 & 0.932 & 2 \\
\bottomrule
\end{tabular}
}
\end{table*}

\section{Holistic Assessment Category Sensitivity}
\label{app:holistic_sensitivity}

The holistic assessment category $S$ contributes one-third of the harmonic-mean weight to ATLAScore and consists of two benchmarks: LongBench-v2 (503 instances) and AA-LCR (100 instances). Because AA-LCR is relatively small in scale, a legitimate concern is whether the overall ranking is overly sensitive to this single component or to the equal one-third weighting of the holistic category. We address this by testing five configurations that vary the composition or weight of $S$, while holding all other design choices fixed.

Table~\ref{tab:holistic_sensitivity} summarizes the results. Removing AA-LCR and using only LongBench-v2 as the holistic score yields $\rho = 0.981$ at 128K with a maximum rank shift of 4; using only AA-LCR yields $\rho = 0.995$ with a maximum shift of 2. Reducing the holistic weight from $\frac{1}{3}$ to $\frac{1}{5}$ (weighted harmonic mean with $w_B{=}0.4, w_C{=}0.4, w_S{=}0.2$) produces $\rho = 0.992$ with a maximum shift of 3. Even the most aggressive variant---dropping $S$ entirely and computing a two-category harmonic mean of $B$ and $C$---still yields $\rho = 0.969$ at 128K and $\rho = 0.992$ at 1M, with a maximum shift of 6 positions.

Two patterns emerge. First, the top-ranked models (Gemini-3.1-Pro-Preview, Claude-Opus-4.6, and GPT-5.5) remain in the top tier across the variants, indicating that their leading positions do not depend on the holistic category. Second, the largest rank perturbations occur for models with pronounced gaps between their holistic scores and their foundational/length-sliced application scores---for example, Kimi-Linear-48B-A3B-Instruct climbs from 24th to 18th when $S$ is dropped, because its very low holistic score (29.4) disproportionately penalizes it under harmonic aggregation.

These results support the inclusion of the holistic category as a stabilizing third signal: it adds diagnostic value by penalizing models that score well on controlled tasks but poorly on realistic long-context reasoning and analysis, while its specific composition (AA-LCR vs.\ LongBench-v2) and weight have limited impact on relative rankings ($\rho \geq 0.981$ for all single-component variants).

\begin{table*}[tp]
\caption{Sensitivity of ATLAScore rankings to the composition and weighting of the holistic assessment category. All variants use harmonic-mean aggregation; only the holistic component or its weight changes. $\rho$ and $\tau$ are computed against the default configuration (equal-weight mean of AA-LCR and LongBench-v2, one-third harmonic weight). Max~$|\Delta r|$ is the largest absolute rank shift for any single model relative to the default ranking.}
\label{tab:holistic_sensitivity}
\centering
\small
\renewcommand{\arraystretch}{1.25}
\resizebox{\textwidth}{!}{
\begin{tabular}{lcccccc}
\toprule
& \multicolumn{3}{c}{@8K-128K} & \multicolumn{3}{c}{@8K-1M} \\
\cmidrule(lr){2-4} \cmidrule(lr){5-7}
Variant & $\rho$ & $\tau$ & Max $|\Delta r|$ & $\rho$ & $\tau$ & Max $|\Delta r|$ \\
\midrule
Default (AA-LCR + LBv2, $\frac{1}{3}$ weight) & 1.000 & 1.000 & 0 & 1.000 & 1.000 & 0 \\
$S$ = LongBench-v2 only & 0.981 & 0.914 & 4 & 0.999 & 0.988 & 1 \\
$S$ = AA-LCR only & 0.995 & 0.957 & 2 & 0.997 & 0.969 & 1 \\
Reduced $S$ weight ($w_B{=}0.4, w_C{=}0.4, w_S{=}0.2$) & 0.992 & 0.951 & 3 & 0.999 & 0.988 & 1 \\
Drop $S$ entirely (2-category H.M.) & 0.969 & 0.877 & 6 & 0.992 & 0.963 & 4 \\
\bottomrule
\end{tabular}
}
\end{table*}

\section{Confidence Interval Propagation}
\label{app:ci}

This appendix derives the confidence interval (CI) propagation chain used to quantify the statistical uncertainty of \atlas scores. The chain proceeds in three stages: subset-level variance estimation, AUC-level linear propagation, and ATLAScore-level delta-method propagation. All intervals are reported at the 95\% confidence level ($z=1.96$). We write $h(q)$ for the 95\% half-width of an estimator $q$ and $\hat v(q)=(h(q)/z)^2$ for its variance estimate. The propagation formulas operate on variances; CIs are obtained only at the end by multiplying the propagated standard error by $z$.

\subsection{Subset-Level Confidence Intervals}

We employ three variance estimators depending on the statistical properties of each component.

\paragraph{CLT base method.}
For components whose instances are approximately independent (OOLong-Synth, GraphWalks, LOFT-Text Retrieval, Helmet-ICL, AMemBench-ACU, LongBench-v2), we apply the central limit theorem directly. Given $n$ instance-level scores $\{x_1,\ldots,x_n\}$ with Bessel-corrected sample standard deviation $\hat\sigma$, the variance and half-width of the subset mean are
\[
\hat v(\bar{x}) = \frac{\hat\sigma^2}{n}, \qquad
h(\bar{x}) = z\sqrt{\hat v(\bar{x})}.
\]

\paragraph{Cluster method.}
For MRCR and AA-LCR, multiple instances share the same base context, inducing within-group correlation that violates the independence assumption. We therefore treat each base context as a cluster $C_k$ and use the cluster-robust variance of the sample mean:
\begin{align*}
\hat v_{\mathrm{cl}}(\bar{x})
&=
\frac{K}{K-1}\frac{1}{n^2}
\sum_{k=1}^{K}
\left(\sum_{i\in C_k}(x_i-\bar{x})\right)^2, \\
h_{\mathrm{cl}}(\bar{x})
&= z\sqrt{\hat v_{\mathrm{cl}}(\bar{x})}.
\end{align*}
where $K$ is the number of base-context clusters. This estimator includes both within-cluster covariance and between-cluster variation while remaining nonnegative.

\paragraph{Weighted combination method.}
LongCodeBench combines binary LongCodeQA and LongSWE scores with instance-count weights. For sub-component $m$, let $\bar{x}_m$, $\hat\sigma_m$, and $n_m$ denote its mean, sample standard deviation, and instance count. With normalized aggregation weight $\lambda_m=w_m/\sum_j w_j$ and independent sub-components, the composite estimator and its variance are
\begin{align*}
\bar{x}_w
&= \sum_m \lambda_m\bar{x}_m, \\
\hat v(\bar{x}_w)
&= \sum_m \lambda_m^2\frac{\hat\sigma_m^2}{n_m}, \\
h(\bar{x}_w)
&= z\sqrt{\hat v(\bar{x}_w)}.
\end{align*}
For instance-count weighting ($w_m=n_m$), this reduces to $\hat v(\bar{x}_w)=\sum_m n_m\hat\sigma_m^2/(\sum_m n_m)^2$.

\subsection{AUC Confidence Interval}

As defined in Section~\ref{sec:scoring}, the AUC for dimension $d$ up to scope $L^\star$ is a normalized trapezoidal integral over the ordered length slices $\{\ell_0,\ell_1,\ldots,\ell_n\}$. Let $y_{d,i}=s_d(\ell_i)/100$ be the slice score on the original 0--1 scale. The trapezoidal rule is linear in the slice scores:
\[
\auc_d(L^\star) = 100\sum_{i=0}^{n}\alpha_i y_{d,i},
\]
where the effective trapezoidal weights are
\[
\alpha_i =
\begin{cases}
\dfrac{\ell_1-\ell_0}{2(\ell_n-\ell_0)}, & i=0,\\[0.6em]
\dfrac{\ell_{i+1}-\ell_{i-1}}{2(\ell_n-\ell_0)}, & 0<i<n,\\[0.6em]
\dfrac{\ell_n-\ell_{n-1}}{2(\ell_n-\ell_0)}, & i=n.
\end{cases}
\]
If the per-slice instance sets are independent and $\hat v(y_{d,i})$ is the subset-level variance on the 0--1 scale, then
\begin{align*}
\hat v\bigl(\auc_d(L^\star)\bigr)
&=
10^4\sum_{i=0}^{n}\alpha_i^2\,\hat v(y_{d,i}), \\
h\bigl(\auc_d(L^\star)\bigr)
&=
z\sqrt{\hat v\bigl(\auc_d(L^\star)\bigr)}.
\end{align*}
Equivalently, if the stored per-slice CI half-width is $h(y_{d,i})$, then $\hat v(y_{d,i})=(h(y_{d,i})/z)^2$.

\subsection{ATLAScore CI via the Delta Method}

The ATLAScore is defined as the harmonic mean of three category aggregates:
\[
H = \frac{3}{\dfrac{1}{B(L^\star)} + \dfrac{1}{C(L^\star)} + \dfrac{1}{S(L^\star)}},
\]
where $B(L^\star)$, $C(L^\star)$, and $S(L^\star)$ are the arithmetic means of the foundational, application, and holistic assessment scores, respectively. Before category aggregation, all component variances are expressed on the 0--100 reporting scale; variances estimated on the 0--1 scale are multiplied by $10^4$. Because $H$ is a nonlinear function of $B$, $C$, and $S$, we use the delta method (first-order Taylor expansion) to approximate its variance.

The partial derivatives of $H$ with respect to each category aggregate are:
\[
\frac{\partial H}{\partial B} = \frac{H^2}{3B^2}, \qquad \frac{\partial H}{\partial C} = \frac{H^2}{3C^2}, \qquad \frac{\partial H}{\partial S} = \frac{H^2}{3S^2}.
\]
Under the working approximation that the category estimates are independent because they aggregate disjoint instance sets, the delta-method variance is
\[
\hat v(H)
\;\approx\;
\frac{H^4}{9}
\left[
\frac{\hat v(B)}{B^4}
+ \frac{\hat v(C)}{C^4}
+ \frac{\hat v(S)}{S^4}
\right],
\]
where
\begin{align*}
\hat v(B)
&= \frac{1}{|\mathcal{D}_{\mathrm{base}}|^2}
\sum_{d \in \mathcal{D}_{\mathrm{base}}} \hat v(\auc_d), \\
\hat v(C)
&= \frac{1}{|\mathcal{D}_{\mathrm{app}}|^2}
\sum_{d \in \mathcal{D}_{\mathrm{app}}} \hat v(\auc_d), \\
\hat v(S)
&= \frac{1}{|\mathcal{D}_{\mathrm{hol}}|^2}
\sum_{d \in \mathcal{D}_{\mathrm{hol}}} \hat v(s_d).
\end{align*}
The 95\% CI for the ATLAScore is $\pm\, h(H)=\pm\, z\sqrt{\hat v(H)}$.

\subsection{Monte Carlo Validation}

To verify the delta-method approximation, we run $10^5$ Monte Carlo simulations for each evaluated model. In each trial, we draw $B^{(t)} \sim \mathcal{N}\bigl(B,\hat v(B)\bigr)$, $C^{(t)} \sim \mathcal{N}\bigl(C,\hat v(C)\bigr)$, and $S^{(t)} \sim \mathcal{N}\bigl(S,\hat v(S)\bigr)$ independently and compute $H^{(t)}$. The empirical 95\% CI is obtained from the 2.5th and 97.5th percentiles of the $\{H^{(t)}\}$ distribution. Across the evaluated models, the ratio of the delta-method CI to the Monte Carlo empirical CI falls in the range $[0.998, 1.004]$, confirming that the first-order approximation introduces negligible error in this regime.

\section{ATLAScore Leaderboards}
\label{app:leaderboard}

Tables~\ref{tab:leaderboard_128k} and~\ref{tab:leaderboard_1m} report the full category-level ATLAScore rankings across all 26 evaluated models.

\begin{table*}[tp]
\caption{Full \atlas leaderboard under ATLAScore@8K-128K. Each cell reports score with 95\% confidence interval.}
\label{tab:leaderboard_128k}
\centering
\scriptsize
\renewcommand{\arraystretch}{1.12}
\resizebox{\textwidth}{!}{
\begin{tabular}{rlcccc}
\toprule
\# & Model & Foundational & Application & Holistic & \textbf{ATLAScore} \\
\midrule
1 & Gemini-3.1-Pro-Preview (high) & 86.36$_{\pm1.96}$ & 74.98$_{\pm2.03}$ & 73.37$_{\pm3.09}$ & \textbf{77.83$_{\pm1.47}$} \\
2 & Claude-Opus-4.6 (max) & 88.96$_{\pm1.70}$ & 72.86$_{\pm2.08}$ & 71.72$_{\pm3.13}$ & \textbf{77.10$_{\pm1.50}$} \\
3 & GPT-5.5 (xhigh) & 89.41$_{\pm1.75}$ & 65.72$_{\pm1.91}$ & 72.49$_{\pm3.11}$ & \textbf{74.63$_{\pm1.43}$} \\
4 & GPT-5.2 (xhigh) & 88.00$_{\pm1.32}$ & 64.26$_{\pm1.70}$ & 72.59$_{\pm4.01}$ & \textbf{73.71$_{\pm1.60}$} \\
5 & GPT-5.4 (xhigh) & 86.03$_{\pm1.95}$ & 66.09$_{\pm1.93}$ & 70.96$_{\pm3.15}$ & \textbf{73.44$_{\pm1.46}$} \\
6 & Gemini-3-Pro-Preview (high) & 80.40$_{\pm1.81}$ & 68.80$_{\pm1.88}$ & 70.69$_{\pm4.35}$ & \textbf{72.96$_{\pm1.77}$} \\
7 & Gemini-3-Flash-Preview (high) & 76.87$_{\pm1.98}$ & 71.73$_{\pm1.83}$ & 67.84$_{\pm4.49}$ & \textbf{71.96$_{\pm1.88}$} \\
8 & DeepSeek-V4-Pro (max) & 82.46$_{\pm2.21}$ & 65.30$_{\pm2.19}$ & 69.06$_{\pm3.24}$ & \textbf{71.56$_{\pm1.56}$} \\
9 & Qwen3.5-397B-A17B (Reasoning) & 85.34$_{\pm1.87}$ & 64.70$_{\pm1.97}$ & 66.40$_{\pm3.30}$ & \textbf{71.03$_{\pm1.55}$} \\
10 & Claude-Opus-4.5 (high) & 69.14$_{\pm1.66}$ & 69.54$_{\pm1.96}$ & 71.00$_{\pm4.22}$ & \textbf{69.88$_{\pm1.62}$} \\
11 & Qwen3.6-Plus (Reasoning) & 84.68$_{\pm1.89}$ & 59.98$_{\pm1.99}$ & 64.87$_{\pm3.30}$ & \textbf{68.34$_{\pm1.55}$} \\
12 & Kimi-K2.6 (Reasoning) & 60.31$_{\pm2.44}$ & 67.43$_{\pm2.09}$ & 66.03$_{\pm3.21}$ & \textbf{64.44$_{\pm1.52}$} \\
13 & Kimi-K2.5 (Reasoning) & 62.37$_{\pm2.42}$ & 65.06$_{\pm2.15}$ & 65.65$_{\pm3.30}$ & \textbf{64.33$_{\pm1.53}$} \\
14 & GLM-5 (Reasoning) & 64.49$_{\pm2.33}$ & 63.23$_{\pm2.17}$ & 65.29$_{\pm3.30}$ & \textbf{64.32$_{\pm1.52}$} \\
15 & GLM-5.1 (Reasoning) & 62.58$_{\pm2.43}$ & 63.58$_{\pm2.14}$ & 65.66$_{\pm3.37}$ & \textbf{63.91$_{\pm1.54}$} \\
16 & Qwen3.5-397B-A17B (Non-reasoning) & 70.93$_{\pm2.45}$ & 59.93$_{\pm1.95}$ & 59.67$_{\pm3.45}$ & \textbf{63.10$_{\pm1.61}$} \\
17 & Kimi-K2.5 (Non-reasoning) & 60.36$_{\pm2.61}$ & 62.05$_{\pm2.16}$ & 57.92$_{\pm3.35}$ & \textbf{60.06$_{\pm1.62}$} \\
18 & GLM-5 (Non-reasoning) & 48.52$_{\pm2.56}$ & 57.40$_{\pm2.06}$ & 44.30$_{\pm3.17}$ & \textbf{49.50$_{\pm1.67}$} \\
19 & Kimi-K2-0905-Instruct & 43.09$_{\pm1.97}$ & 53.90$_{\pm2.08}$ & 47.10$_{\pm4.61}$ & \textbf{47.63$_{\pm1.85}$} \\
20 & Qwen3-235B-A22B-2507-Instruct & 50.17$_{\pm1.94}$ & 52.38$_{\pm1.94}$ & 39.52$_{\pm4.14}$ & \textbf{46.64$_{\pm2.07}$} \\
21 & DeepSeek-V3.1 (Non-reasoning) & 50.30$_{\pm1.89}$ & 41.46$_{\pm1.96}$ & 45.80$_{\pm4.78}$ & \textbf{45.57$_{\pm1.84}$} \\
22 & Qwen3-Next-80B-A3B-Instruct & 41.59$_{\pm2.07}$ & 45.86$_{\pm2.15}$ & 46.56$_{\pm4.71}$ & \textbf{44.56$_{\pm1.78}$} \\
23 & DeepSeek-V3.2 (Non-reasoning) & 51.70$_{\pm2.02}$ & 37.53$_{\pm1.96}$ & 43.54$_{\pm4.25}$ & \textbf{43.51$_{\pm1.73}$} \\
24 & Kimi-Linear-48B-A3B-Instruct & 63.47$_{\pm1.98}$ & 47.75$_{\pm2.07}$ & 29.40$_{\pm3.95}$ & \textbf{42.43$_{\pm2.81}$} \\
25 & GLM-4.7 (Non-reasoning) & 43.37$_{\pm1.91}$ & 45.77$_{\pm2.04}$ & 32.93$_{\pm4.15}$ & \textbf{39.85$_{\pm2.16}$} \\
26 & Qwen3-30B-A3B-2507-Instruct & 32.18$_{\pm2.07}$ & 36.58$_{\pm2.08}$ & 31.72$_{\pm4.35}$ & \textbf{33.36$_{\pm1.86}$} \\
\bottomrule
\end{tabular}
}
\end{table*}

\begin{table*}[tp]
\caption{Full \atlas leaderboard under ATLAScore@8K-1M. Each cell reports score with 95\% confidence interval.}
\label{tab:leaderboard_1m}
\centering
\scriptsize
\renewcommand{\arraystretch}{1.12}
\resizebox{\textwidth}{!}{
\begin{tabular}{rlcccc}
\toprule
\# & Model & Foundational & Application & Holistic & \textbf{ATLAScore} \\
\midrule
1 & Claude-Opus-4.6 (max) & 77.19$_{\pm2.12}$ & 63.99$_{\pm1.99}$ & 71.72$_{\pm3.13}$ & \textbf{70.55$_{\pm1.42}$} \\
2 & Gemini-3.1-Pro-Preview (high) & 66.25$_{\pm2.36}$ & 66.41$_{\pm2.01}$ & 73.37$_{\pm3.09}$ & \textbf{68.52$_{\pm1.42}$} \\
3 & GPT-5.5 (xhigh) & 78.64$_{\pm2.22}$ & 56.32$_{\pm1.78}$ & 72.49$_{\pm3.11}$ & \textbf{67.77$_{\pm1.36}$} \\
4 & GPT-5.4 (xhigh) & 65.37$_{\pm2.47}$ & 55.38$_{\pm1.89}$ & 70.96$_{\pm3.15}$ & \textbf{63.23$_{\pm1.40}$} \\
5 & Gemini-3-Flash-Preview (high) & 61.53$_{\pm1.99}$ & 60.50$_{\pm1.71}$ & 67.84$_{\pm4.49}$ & \textbf{63.13$_{\pm1.60}$} \\
6 & Gemini-3-Pro-Preview (high) & 58.43$_{\pm1.94}$ & 60.01$_{\pm1.82}$ & 70.69$_{\pm4.35}$ & \textbf{62.60$_{\pm1.51}$} \\
7 & DeepSeek-V4-Pro (max) & 56.62$_{\pm2.64}$ & 53.74$_{\pm2.18}$ & 69.06$_{\pm3.24}$ & \textbf{59.11$_{\pm1.52}$} \\
8 & GPT-5.2 (xhigh) & 58.70$_{\pm1.86}$ & 48.22$_{\pm1.34}$ & 72.59$_{\pm4.01}$ & \textbf{58.20$_{\pm1.24}$} \\
9 & Qwen3.6-Plus (Reasoning) & 58.59$_{\pm2.30}$ & 45.28$_{\pm1.73}$ & 64.87$_{\pm3.30}$ & \textbf{54.98$_{\pm1.34}$} \\
10 & Qwen3.5-397B-A17B (Reasoning) & 55.87$_{\pm2.51}$ & 45.13$_{\pm1.44}$ & 66.40$_{\pm3.30}$ & \textbf{54.43$_{\pm1.29}$} \\
11 & Claude-Opus-4.5 (high) & 46.38$_{\pm1.68}$ & 45.02$_{\pm1.53}$ & 71.00$_{\pm4.22}$ & \textbf{51.85$_{\pm1.23}$} \\
12 & Qwen3.5-397B-A17B (Non-reasoning) & 50.91$_{\pm2.53}$ & 43.72$_{\pm1.39}$ & 59.67$_{\pm3.45}$ & \textbf{50.61$_{\pm1.33}$} \\
13 & Kimi-K2.6 (Reasoning) & 43.31$_{\pm2.43}$ & 47.14$_{\pm1.67}$ & 66.03$_{\pm3.21}$ & \textbf{50.47$_{\pm1.42}$} \\
14 & Kimi-K2.5 (Reasoning) & 45.76$_{\pm2.39}$ & 37.60$_{\pm1.86}$ & 65.65$_{\pm3.30}$ & \textbf{47.11$_{\pm1.41}$} \\
15 & GLM-5.1 (Reasoning) & 37.79$_{\pm2.29}$ & 43.40$_{\pm1.46}$ & 65.66$_{\pm3.37}$ & \textbf{46.35$_{\pm1.39}$} \\
16 & GLM-5 (Reasoning) & 35.29$_{\pm2.25}$ & 41.49$_{\pm1.53}$ & 65.29$_{\pm3.30}$ & \textbf{44.28$_{\pm1.41}$} \\
17 & Kimi-K2.5 (Non-reasoning) & 44.30$_{\pm2.47}$ & 29.36$_{\pm1.81}$ & 57.92$_{\pm3.35}$ & \textbf{40.60$_{\pm1.45}$} \\
18 & GLM-5 (Non-reasoning) & 31.12$_{\pm2.26}$ & 39.01$_{\pm1.35}$ & 44.30$_{\pm3.17}$ & \textbf{37.34$_{\pm1.38}$} \\
19 & Qwen3-235B-A22B-2507-Instruct & 29.53$_{\pm1.56}$ & 37.82$_{\pm1.43}$ & 39.52$_{\pm4.14}$ & \textbf{35.04$_{\pm1.37}$} \\
20 & Kimi-K2-0905-Instruct & 27.74$_{\pm1.45}$ & 32.00$_{\pm1.72}$ & 47.10$_{\pm4.61}$ & \textbf{33.89$_{\pm1.25}$} \\
21 & Kimi-Linear-48B-A3B-Instruct & 45.51$_{\pm2.02}$ & 28.87$_{\pm1.70}$ & 29.40$_{\pm3.95}$ & \textbf{33.10$_{\pm1.86}$} \\
22 & Qwen3-Next-80B-A3B-Instruct & 22.19$_{\pm1.24}$ & 26.50$_{\pm1.58}$ & 46.56$_{\pm4.71}$ & \textbf{28.77$_{\pm1.11}$} \\
23 & DeepSeek-V3.1 (Non-reasoning) & 32.78$_{\pm1.59}$ & 18.85$_{\pm1.16}$ & 45.80$_{\pm4.78}$ & \textbf{28.47$_{\pm1.15}$} \\
24 & DeepSeek-V3.2 (Non-reasoning) & 33.24$_{\pm1.63}$ & 17.98$_{\pm1.19}$ & 43.54$_{\pm4.25}$ & \textbf{27.61$_{\pm1.16}$} \\
25 & Qwen3-30B-A3B-2507-Instruct & 20.39$_{\pm1.41}$ & 23.93$_{\pm1.56}$ & 31.72$_{\pm4.35}$ & \textbf{24.52$_{\pm1.23}$} \\
26 & GLM-4.7 (Non-reasoning) & 16.88$_{\pm0.92}$ & 9.92$_{\pm0.90}$ & 32.93$_{\pm4.15}$ & \textbf{15.76$_{\pm0.86}$} \\
\bottomrule
\end{tabular}
}
\end{table*}

\section{\atlaslite: Single-Slice Screening at 128K}
\label{app:atlas_lite}

The full \atlas protocol evaluates every model over eight geometrically spaced length slices from 8K to 1M tokens. While this design provides a detailed degradation curve, it also demands substantial token throughput: approximately 1,711M tokens per model, of which the 1M slice alone accounts for roughly 45\%. Motivated by prior work on low-cost long-context evaluation~\citep{huang2025minilongbench}, this appendix investigates whether a single carefully chosen length slice can approximate the full ranking at dramatically lower cost, and characterizes the cost--fidelity Pareto frontier across several candidate reduction schemes.

\subsection{Why 128K Is the Natural Single-Slice Choice}

The 128K slice occupies a unique position in the current model landscape: it is the longest context length at which most evaluated models still operate within or near their pretraining window and exhibit broadly stable performance. Beyond 128K, many models experience sharp capability cliffs. For example, the per-slice ATLAScore of Claude-Opus-4.5 drops from 66.4 at 128K to 54.4 at 256K (an 18\% relative decline), and GLM-4.7 drops from 30.1 at 128K to 7.0 at 256K (a 77\% decline). These cliff effects make post-128K slices informative for stress testing but unreliable as proxies for general capability ranking, because they conflate genuine long-context weakness with context-window overflow.

Selecting 128K as the sole evaluation point therefore captures the most discriminative performance regime---where model quality differences are driven by architecture and training rather than by whether a model's context window physically accommodates the input---while avoiding the noise introduced by cliff effects.

\subsection{Cost--Fidelity Analysis}

Table~\ref{tab:atlas_lite_schemes} compares multiple slice-subset schemes by evaluating all 26 models (15 reasoning, 11 non-reasoning) on the full \atlas benchmark and computing the Spearman rank correlation ($\rho$) between each reduced scheme's ranking and the full eight-slice ranking. Token cost is measured as total input tokens consumed per model; the fixed cost of single-value subsets (LongBench-v2 and AA-LCR, approximately 145M tokens) is included in all schemes.

\begin{table*}[tp]
\caption{Cost--fidelity trade-off for candidate slice-reduction schemes. Relative cost is the fraction of the full eight-slice token budget. $\rho$ denotes Spearman rank correlation against the full ranking. Max~$|\Delta r|$ is the largest absolute rank shift for any single model relative to the full eight-slice ranking. Efficiency is defined as $\rho\,/\,$relative cost.}
\label{tab:atlas_lite_schemes}
\centering
\small
\renewcommand{\arraystretch}{1.25}
\resizebox{\textwidth}{!}{
\begin{tabular}{lrrrrr}
\toprule
Scheme & Rel.\ cost & $\rho$ & Max $|\Delta r|$ & $\geq$2 shifts & Efficiency \\
\midrule
Full 8 slices & 100.0\% & 1.000 & 0 & 0 & 1.00 \\
7 slices (drop 1M) & 54.5\% & 0.997 & 1 & 0 & 1.83 \\
3 pts: 8K+128K+1M & 60.3\% & 0.996 & 2 & 1 & 1.65 \\
6 slices (drop 512K+1M) & 31.7\% & 0.988 & 4 & 2 & 3.12 \\
5 slices: 8K--128K & 19.9\% & 0.976 & 4 & 7 & 4.90 \\
\textbf{128K only (\atlaslite)} & \textbf{14.5\%} & \textbf{0.977} & \textbf{4} & \textbf{5} & \textbf{6.74} \\
256K only & 20.2\% & 0.987 & 3 & 6 & 4.89 \\
8K + 128K & 14.8\% & 0.969 & 4 & 11 & 6.55 \\
\bottomrule
\end{tabular}
}
\end{table*}

Two observations stand out. First, the 128K-only scheme achieves the highest efficiency (6.74) among all candidates while retaining high rank correlation ($\rho = 0.977$). Second, the 256K-only scheme reaches higher rank fidelity ($\rho=0.987$) but costs 40\% more and is less attractive for rapid screening because it already reflects post-128K cliff effects for several models.

\subsection{Pareto Frontier}

Among all candidate schemes, five are Pareto-optimal---no other scheme achieves higher $\rho$ at equal or lower cost:

\begin{enumerate}[leftmargin=1.2em]
\item \textbf{128K only} (\atlaslite): 14.5\% cost, $\rho = 0.977$. Highest efficiency.
\item \textbf{256K only}: 20.2\% cost, $\rho = 0.987$. Higher fidelity at a modestly larger cost.
\item \textbf{6 slices} (drop 512K+1M): 31.7\% cost, $\rho = 0.988$. Slightly higher fidelity before the ultra-long slices.
\item \textbf{7 slices} (drop 1M): 54.5\% cost, $\rho = 0.997$. Near-perfect fidelity without the expensive 1M slice.
\item \textbf{Full 8 slices}: 100\% cost, $\rho = 1.0$. The reference protocol.
\end{enumerate}

The marginal cost of moving from \atlaslite ($\rho = 0.977$) to the 7-slice variant ($\rho = 0.997$) is a 3.8$\times$ increase in token budget for a 0.020 improvement in rank correlation. All other schemes are dominated: they achieve lower $\rho$ at equal or higher cost than one of the Pareto-optimal options.

\subsection{Rank Stability Under \atlaslite}

Of the 26 evaluated models, 21 shift by at most 1 rank position under \atlaslite; 2 models shift by exactly 2 positions, and 3 models shift by 4 positions. The largest shifts occur for models whose post-128K degradation differs sharply from their 128K performance, including GPT-5.2, Gemini-3-Flash, and Qwen3.5-397B-A17B (Reasoning). This stability is sufficient for first-pass screening, but it also reinforces the main result: a single 128K slice captures much of the full AUC-based ranking while still missing the post-128K degradation profile.

\subsection{Limitations and Recommended Usage}

The primary limitation of \atlaslite is that it cannot detect performance cliffs that emerge at 256K, 512K, or 1M. For teams whose deployment scenario involves inputs longer than 128K, the full protocol or the 7-slice variant remains necessary. \atlaslite is best suited for three operational scenarios: (1)~rapid screening of newly released models, (2)~regression detection during iterative model development, and (3)~initial triage to narrow the candidate set before committing to a full evaluation. In all cases, the full \atlas protocol remains the recommended standard for official benchmarking and public reporting.

\section{Length Degradation Analysis}
\label{app:decay}

This appendix provides two complementary views of how model performance degrades as context length increases: a decay heatmap that summarizes the degradation ratio between the 128K and 1M reporting scopes, and per-slice length curves that reveal the trajectory of degradation.

\subsection{Capability Decay Heatmap}

We define the \emph{capability decay rate} for model $m$ on dimension $d$ as
\[
\mathrm{decay}(m, d) = \frac{\auc_d^{\text{@128K}}(m) - \auc_d^{\text{@1M}}(m)}{\auc_d^{\text{@128K}}(m)},
\]
where $\auc_d^{\text{@128K}}$ and $\auc_d^{\text{@1M}}$ denote the cumulative AUC scores up to 128K and 1M, respectively. A decay rate of 0\% indicates no degradation; higher values indicate greater loss. Negative values (rare) indicate slight improvement when longer contexts are included.

\begin{figure*}[tp]
\centering
\includegraphics[width=\textwidth]{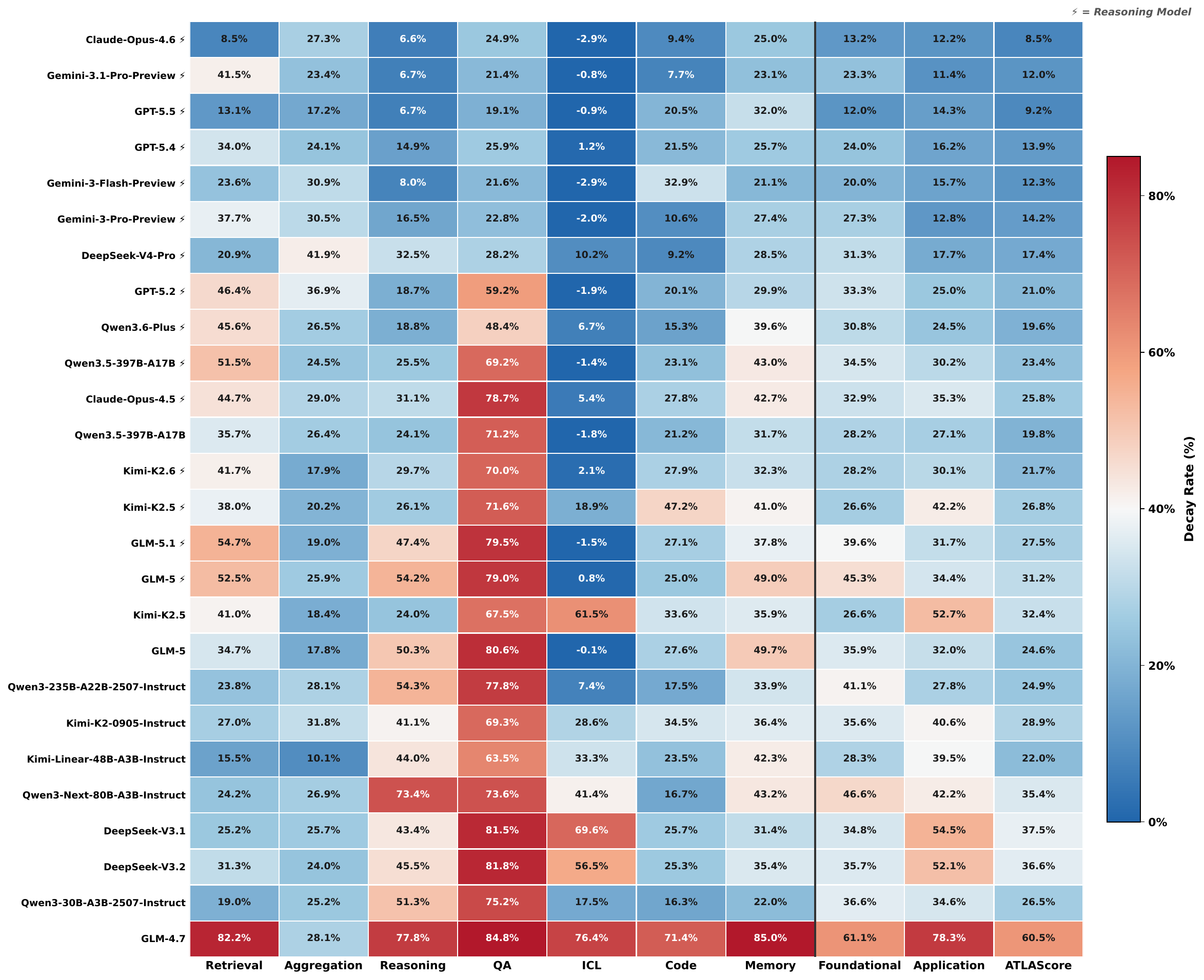}
\caption{Full capability decay heatmap across all 26 evaluated models, ordered by ATLAScore@8K-1M. Blue cells indicate low decay and red cells indicate severe degradation. This is the complete version of the representative-model view shown in Figure~\ref{fig:decay_heatmap}.}
\label{fig:decay_heatmap_full}
\end{figure*}

Figure~\ref{fig:decay_heatmap_full} visualizes this quantity across all 26 models, seven length-sliced dimensions, and three aggregate scores, revealing several patterns.

\paragraph{Overall decay spectrum.}
ATLAScore decay ranges from 8.5\% (Claude-Opus-4.6) to 60.5\% (GLM-4.7). As documented in Section~\ref{sec:decay}, reasoning-labeled configurations cluster in the low-decay region, while non-reasoning models span the full range.

\paragraph{Dimension-specific fragility.}
Retrieval and QA are the most decay-prone dimensions, with multiple models exceeding 70\% decay. For instance, GLM-4.7 loses 82.2\% on Retrieval and 84.8\% on QA. By contrast, Code understanding exhibits uniformly low decay across nearly all models. This pattern suggests that retrieval-intensive tasks are disproportionately affected by context extension, while code understanding tasks, which rely more on local structure, are relatively insulated.

\paragraph{Heterogeneous within-model profiles.}
The heatmap reveals extreme within-model variation that aggregate decay numbers would obscure. GLM-5, for example, shows 80.6\% decay on QA but essentially no degradation on ICL ($-0.1\%$), a spread exceeding 80 percentage points within a single model. Similarly, Claude-4.5 shows 31.1\% decay on Reasoning but 78.7\% on QA. This reinforces the main conclusion that models that appear broadly capable at 128K may have severe dimension-specific blind spots at longer contexts.

\paragraph{Implications for model development.}
The decay heatmap provides actionable diagnostic information beyond what aggregate degradation numbers capture. Model developers can identify which specific capabilities degrade most under context extension and prioritize targeted improvements. The correlation between reasoning-model status and low overall decay is a useful hypothesis generator for future work, but it should be interpreted as a deployed-configuration association rather than evidence for a specific causal mechanism.

\subsection{Length Degradation Curves}
\label{app:length_curves}

The decay heatmap summarizes degradation as a single ratio between two reporting scopes. The following per-slice ATLAScore curves complement that view by revealing the \emph{trajectory} of degradation from 8K to 1M tokens.

\begin{figure*}[tp]
\centering
\includegraphics[width=\textwidth]{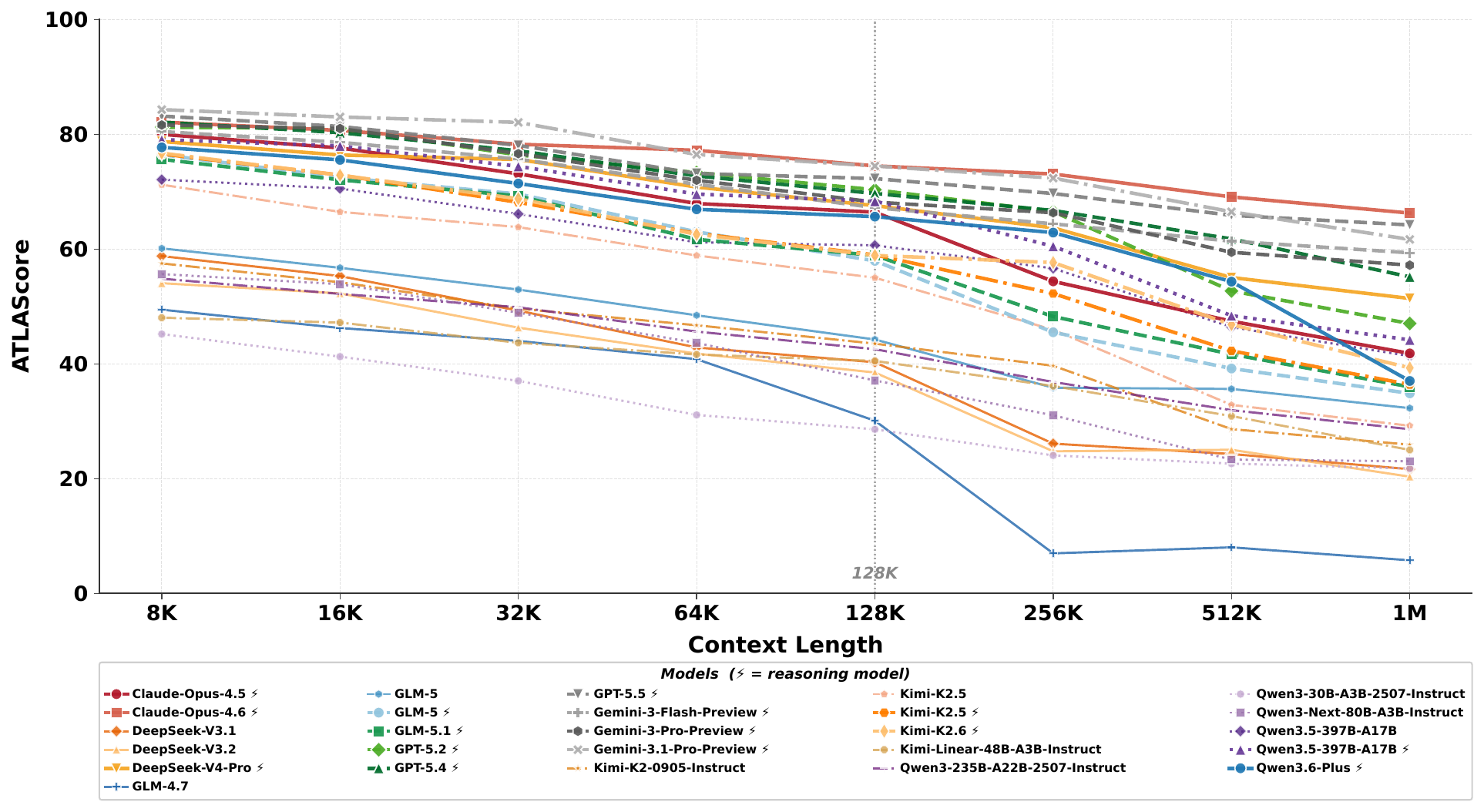}
\caption{Per-slice ATLAScore across all 26 evaluated models from 8K to 1M tokens. Solid lines denote reasoning models; dashed lines denote non-reasoning models. The vertical dotted line marks the 128K boundary that separates the two reporting scopes.}
\label{fig:length_curve_full}
\end{figure*}

Figure~\ref{fig:length_curve_full} plots the per-slice ATLAScore for all 26 models. Several patterns emerge beyond what the two-scope comparison in the main text can capture.

\paragraph{Cliff effects vs.\ gradual degradation.}
Most models exhibit roughly linear degradation on the log-length axis from 8K through 128K, but the post-128K regime bifurcates sharply. A first group---including GPT-5.5, Claude-Opus-4.6, and Gemini-3-Flash-Preview---continues to degrade gradually, losing only 5--15 points between 128K and 1M. A second group experiences abrupt cliffs: GLM-4.7 drops from 30.1 at 128K to 5.8 at 1M, including a fall to 7.0 at 256K, and Claude-Opus-4.5 loses 12.0 points between 128K and 256K alone. These cliff effects are invisible in a two-scope summary but are immediately apparent in the length curve.

\paragraph{Reasoning vs.\ non-reasoning trajectories.}
Reasoning models (solid lines) start higher on average and remain higher at every length slice. Their relative degradation is milder, even when absolute 128K-to-1M drops can be similar to those of non-reasoning models. Non-reasoning models exhibit lower starting points and the widest spread between 256K and 1M. This visual pattern corroborates the statistical observation in the decay heatmap above that reasoning-labeled deployed configurations exhibit systematically lower decay rates.

\section{Rank Migration from 128K to 1M}
\label{app:rank_migration}

The main text reports that 7 of 26 models shift by two or more rank positions when the reporting scope extends from ATLAScore@8K-128K to ATLAScore@8K-1M. This appendix visualizes the full rank migration pattern.

\begin{figure*}[tp]
\centering
\includegraphics[width=\textwidth]{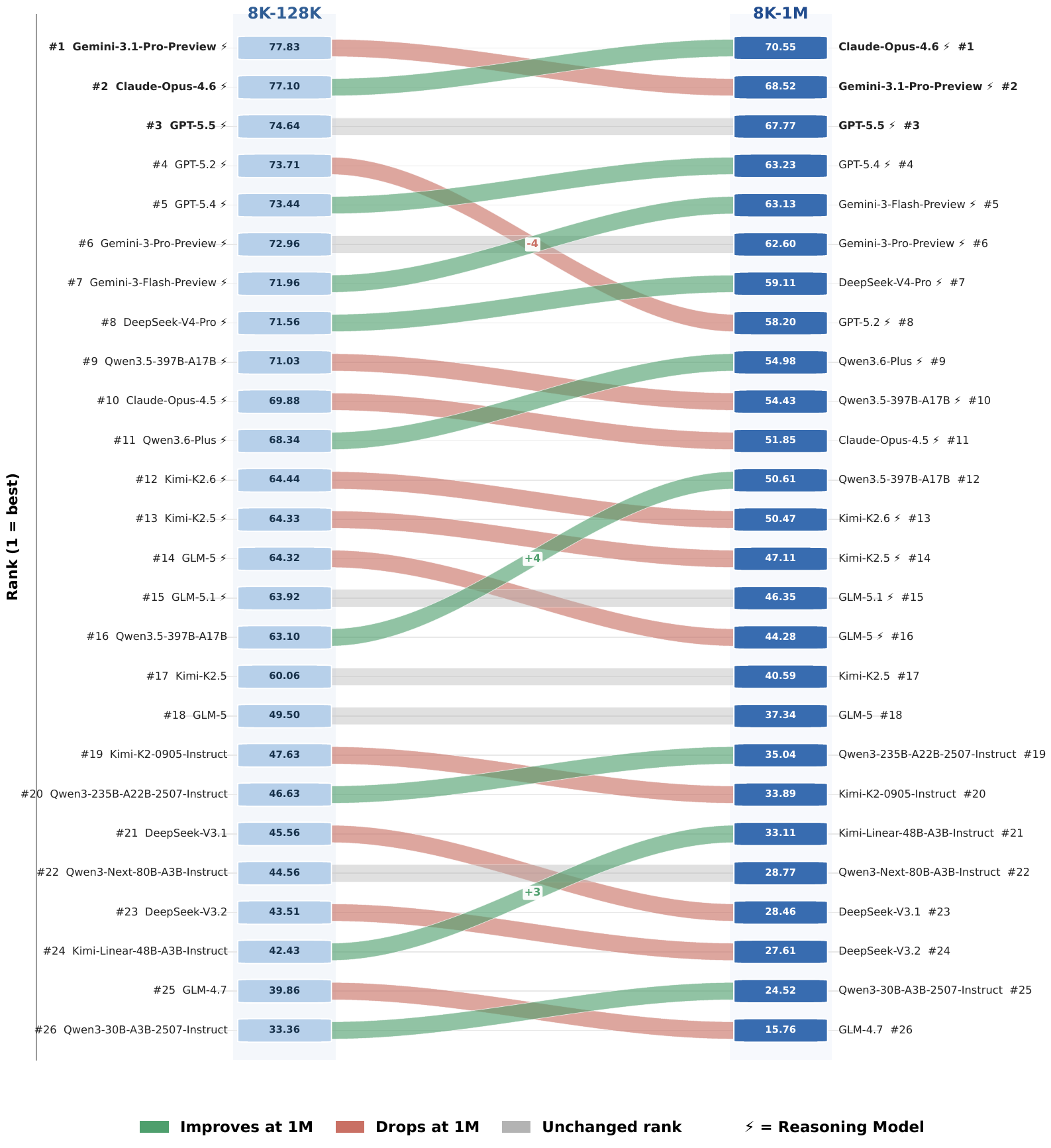}
\caption{Sankey-style rank-flow diagram showing how all 26 models migrate from ATLAScore@8K-128K (left) to ATLAScore@8K-1M (right). Each row is a rank, node labels report the corresponding ATLAScore, and ribbons connect the same model across the two scopes. Green ribbons indicate rank improvements at 1M, red ribbons indicate rank drops, and gray ribbons indicate unchanged ranks. Signed annotations mark larger shifts, with $\Delta r = r_{128\mathrm{K}} - r_{1\mathrm{M}}$, so positive values indicate improvement at 1M.}
\label{fig:rank_sankey}
\end{figure*}

Figure~\ref{fig:rank_sankey} presents a Sankey-style view of rank migration between the two reporting scopes. The top tier is relatively stable: the same three models remain in the top three, but Gemini-3.1-Pro-Preview leads at 128K, Claude-Opus-4.6 leads at 1M, and GPT-5.5 remains third in both scopes. The middle tier is where the largest disruptions occur. GPT-5.2 drops from rank 4 to rank 8, while Gemini-3-Flash-Preview rises from rank 7 to rank 5 and Gemini-3-Pro-Preview remains unchanged at rank 6. In the lower tier, Kimi-Linear-48B-A3B-Instruct rises from rank 24 to rank 21, while DeepSeek-V3.1 drops from rank 21 to rank 23. Because every model has a lower absolute ATLAScore at 1M than at 128K, these flows should be read as relative robustness differences rather than absolute performance gains.

\section{Capability Profile Case Studies}
\label{app:case_studies}

\begin{figure*}[tp]
\centering
\includegraphics[width=\textwidth]{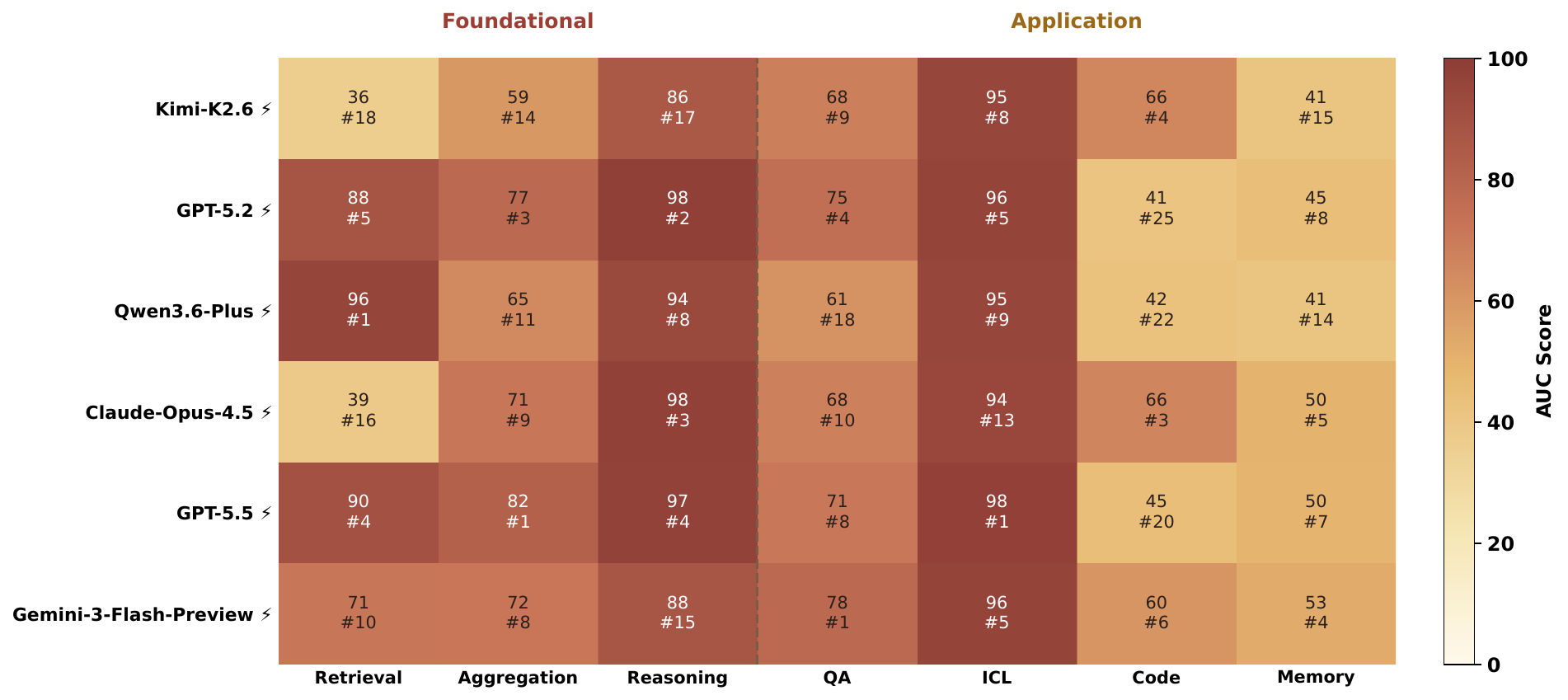}
\caption{Per-dimension score and rank breakdown for the six models with the largest foundational--application rank discrepancies at 128K. The dashed vertical line separates foundational components from length-sliced application components. Each cell shows the AUC score and the component rank.}
\label{fig:case_study}
\end{figure*}

The layer discrepancy in Section~\ref{sec:layer_validation} is best understood at the dimension level, where it becomes a practical model-selection warning. Figure~\ref{fig:case_study} decomposes the most rank-discrepant models into per-dimension scores and ranks, showing that different models reach similar aggregate scores through very different capability profiles.

Kimi-K2.6 is not a strong foundational reader by \atlas standards: its Retrieval score is only 35.58 at 128K. Yet it performs well on length-sliced application tasks because it is strong on ICL (95.23) and Code (65.57). GPT-5.2 shows the reverse pattern: it is near the top on all three foundational dimensions at 128K, but its Code score is only 41.41 and its Memory score is 44.65, pulling down the length-sliced application aggregate. Qwen3.6-Plus is similarly foundational-heavy: it ranks first on Retrieval but only 22nd on Code and 18th on QA. These cases would be hard to diagnose from either a single task or a single aggregate leaderboard.

\section{Capability Radar Charts}
\label{app:radar}

The leaderboard tables (Appendix~\ref{app:leaderboard}) and case-study heatmap (Appendix~\ref{app:case_studies}) provide numerical views of per-dimension performance. This appendix supplements those with radar charts that visualize the full capability profile of all 26 models at both reporting scopes, making it easier to identify characteristic shapes---balanced profiles, single-dimension spikes, and systematic weaknesses---at a glance.

\begin{figure*}[tp]
\centering
\includegraphics[width=\textwidth]{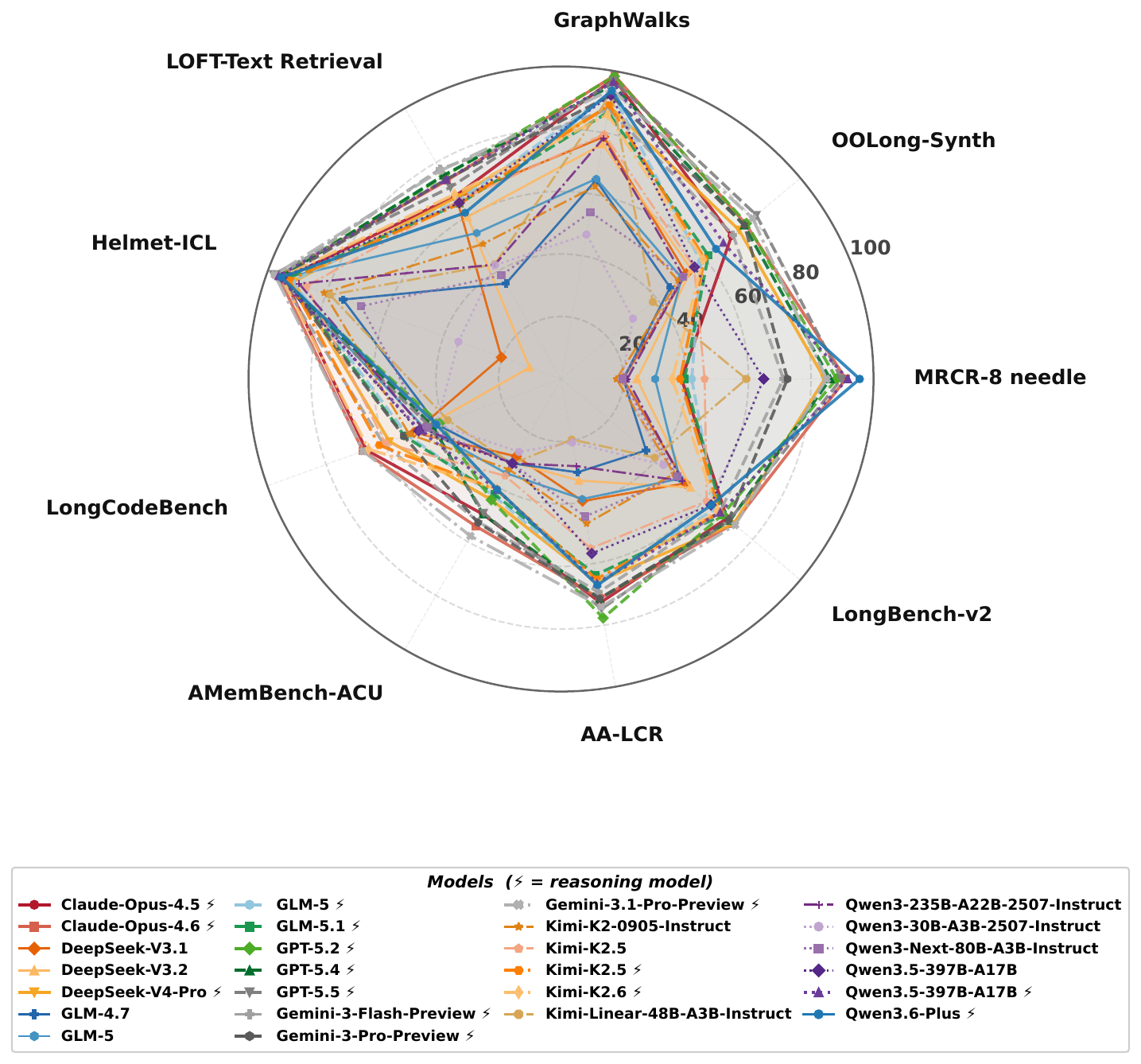}
\caption{Capability radar chart at the 128K reporting scope for all 26 evaluated models. Each axis represents one of the nine \atlas components (seven length-sliced dimensions plus two holistic assessment benchmarks). Reasoning models are shown with solid lines; non-reasoning models with dashed lines. Some models form large, balanced polygons (e.g., Gemini-3.1-Pro-Preview and Claude-Opus-4.6), while others show pronounced spikes or dents on specific axes (e.g., Kimi-K2.6 with weaker Retrieval but strong ICL and Code).}
\label{fig:radar_128k}
\end{figure*}

\begin{figure*}[tp]
\centering
\includegraphics[width=\textwidth]{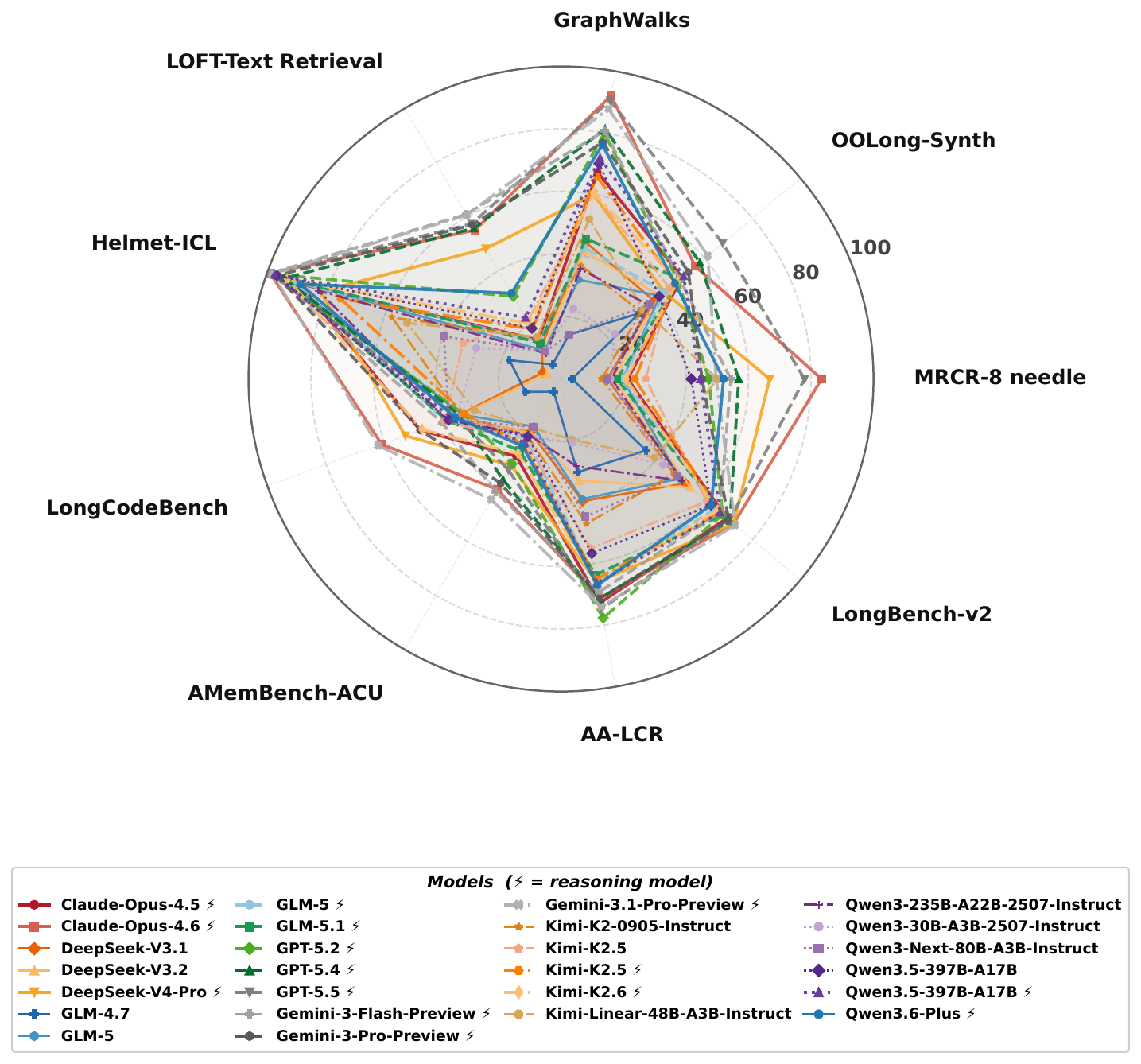}
\caption{Capability radar chart at the 1M reporting scope. Compared to the 128K chart (Figure~\ref{fig:radar_128k}), the overall polygons contract substantially, and the contraction is non-uniform across axes: Retrieval and QA axes shrink the most, while ICL and Code axes retain more of their 128K extent. The shape distortion between the two figures directly reflects the dimension-specific decay analyzed in Appendix~\ref{app:decay}.}
\label{fig:radar_1m}
\end{figure*}

Several observations emerge from comparing the two radar charts.

\paragraph{Profile diversity increases at 1M.}
At 128K (Figure~\ref{fig:radar_128k}), frontier models form roughly similar polygons with high scores on most axes, differing mainly in Code and Memory. At 1M (Figure~\ref{fig:radar_1m}), the profiles diverge more sharply: Claude-Opus-4.6 and Gemini-3.1-Pro-Preview retain the largest profiles, while GPT-5.2's polygon develops a deep indentation on the Retrieval and QA axes. This visual divergence corresponds to the rank reshuffling documented in Section~\ref{sec:experiments}.

\paragraph{Dimension-specific collapse is visually salient.}
The radar format makes it immediately apparent when a model's profile is dominated by a single weakness. GLM-4.7, for instance, contracts to a thin sliver at 1M, with nearly all axes near zero except modest residual scores on Aggregation and Code. Models like GLM-5.1 show a different asymmetry: high ICL and holistic scores coexist with weak Retrieval and QA, forming a lopsided polygon that would be difficult to characterize with a single aggregate score.

\paragraph{Practical use.}
Radar charts are most useful for pairwise or small-group model comparison rather than for reading exact scores. A practitioner choosing between Claude-Opus-4.6 and Gemini-3.1-Pro-Preview at 1M can see at a glance that Claude-Opus-4.6 has a slight edge on Retrieval and Reasoning while Gemini-3.1-Pro-Preview is stronger on QA and LongBench-v2. Such trade-offs are difficult to extract from leaderboard tables but are immediately visible in the radar overlay.

\end{document}